\documentclass{article}
\usepackage[utf8]{inputenc}
\usepackage{hyperref}
\usepackage{amsmath,amsfonts,amsthm,amssymb}
\usepackage{nicefrac}
\usepackage{geometry}
\usepackage{graphicx}
\usepackage{multirow}
\usepackage{glossaries-prefix}
\usepackage{pgfplots}
\usepackage{authblk}
\usepackage{caption}
\usepackage{subcaption}
\usepackage{booktabs}
\usepackage{todonotes}
\usepackage{cleveref}
\usepackage{algorithm2e}
\newtheorem{theorem}{Theorem}
\theoremstyle{definition}
\newtheorem{example}{Example}
\newtheorem{remark}{Remark}

\newcommand{\de}{\mathrm{d}}
\newcommand{\R}{\mathbb{R}}

\title{\textbf{Pseudo-Hamiltonian neural networks for learning partial differential equations}}
\author{Sølve Eidnes and Kjetil Olsen Lye}
\date{January 2, 2024}
\affil{\small{Department of Mathematics and Cybernetics, SINTEF Digital, 0373 Oslo, Norway}\\
\texttt{\{solve.eidnes, kjetil.olsen.lye\}@sintef.no}}

\begin{document}

\maketitle

\begin{abstract}
    Pseudo-Hamiltonian neural networks (PHNN) were recently introduced for learning dynamical systems that can be modelled by ordinary differential equations. In this paper, we extend the method to partial differential equations. The resulting model is comprised of up to three neural networks, modelling terms representing conservation, dissipation and external forces, and discrete convolution operators that can either be learned or be given as input. We demonstrate numerically the superior performance of PHNN compared to a baseline model that models the full dynamics by a single neural network. Moreover, since the PHNN model consists of three parts with different physical interpretations, these can be studied separately to gain insight into the system, and the learned model is applicable also if external forces are removed or changed.
\end{abstract}

\section{Introduction}\label{sec:intro}
The field called physics-informed machine learning combines the strengths of physics-based models and data-driven techniques to achieve a deeper understanding and improved predictive capabilities for complex physical systems \cite{Karniadakis2021physics, Willard2022integrating}. The rapidly growing interest in this interdisciplinary approach is largely motivated by the increasing capabilities of computers to store and process large quantities of data, along with the decreasing costs of sensors and computers that capture and handle data from physical systems. Machine learning for differential equations can broadly be divided into two categories: the forward problem, which involves predicting future states from an initial state, and the inverse problem, which entails learning a system or parts of it from data. A wealth of recent literature exists on machine learning for the forward problem in the context of partial differential equations (PDEs). The proposed methods include neural-network-based substitutes for numerical solvers \cite{E2017deep, E2018deep, Sirignano2018dgm, Raissi2019physics}, but also methods that can aid the solution process, e.g.\ by optimizing the discretization to be used in a solver \cite{Bar2019learning}. The focus of this paper is on the inverse problem, and much of the foundation for our proposed model can be found in recent advances in learning neural network models for ordinary differential equations (ODEs). Specifically, we build on recent works on models that incorporate Hamiltonian mechanics and related structures that underlie the physical systems we seek to model.

Greydanus et al.\ introduced Hamiltonian neural networks (HNN) in \cite{Greydanus2019hamiltonian}, for learning finite-dimensional Hamiltonian systems from data. They assume that the data $q \in \mathbb{R}^n$, $p \in \mathbb{R}^n$ is obtained from a canonical Hamiltonian system
\begin{equation*}
\begin{pmatrix}
    \dot{q} \\
    \dot{p}
\end{pmatrix}
=
\begin{pmatrix}
    0 & I_n \\
    -I_n & 0
\end{pmatrix}
\begin{pmatrix}
    \frac{\partial H}{\partial q} \\
    \frac{\partial H}{\partial p}
\end{pmatrix},
\end{equation*}
and aim to learn a neural network model $\hat{H}_\theta$ of the Hamiltonian $H : \mathbb{R}^n \times \mathbb{R}^n \rightarrow \mathbb{R}$. This approach has since been further explored and expanded in a number of directions, which include considering control input \cite{Zhong2020symplectic}, dissipative systems \cite{Zhong2020dissipative, Greydanus2022dissipative}, constrained systems \cite{Finzi2020simplifying, Celledoni2022learning}, port-Hamiltonian systems \cite{Desai2021port, Duong2021hamiltonian, Duong2022adaptive}, and metriplectic system formulations \cite{Lee2021machine, Hernandez2023port}. A similar approach considering a Lagrangian formulation instead of a Hamiltonian is presented in \cite{Cranmer2020lagrangian}. Modifications of the models that focus on improved training from sparse and noisy data have been proposed in \cite{Chen2019symplectic, Jin2020sympnets, David2021symplectic, Chen2021neural}.

In \cite{Eidnes2023pseudo}, we proposed pseudo-Hamiltonian neural networks (PHNN). These can learn what we call pseudo-Hamiltonian systems, which generalizes Hamiltonian systems first to any invariant-preserving system and further allows for dissipation and external forces acting on the system. Thus we consider the formulation
\begin{equation}\label{eq:noncangen}
    \dot{x} = (S(x) - R(x)) \nabla H(x) + f(x,t), \qquad x \in \mathbb{R}^d,
\end{equation}
where $S(x) = -S(x)^T$, and $y^TR(x)y \geq 0$ for all $y$. That is, $S(x) \in \mathbb{R}^{d\times d}$ can be any skew-symmetric matrix and $R(x) \in \mathbb{R}^{d\times d}$ can be any positive semi-definite matrix. Since we put no restrictions on the external forces, a pseudo-Hamiltonian formulation can in principle be obtained for any first-order ODE, which in turn can be obtained from any arbitrary-order ODE by a variable transformation. 
The formulation makes it possible to learn models that can be separated into internal dynamics and the external forces, i.e.\ $(\hat{S}_\theta(x) - \hat{R}_\theta(x)) \nabla \hat{H}_\theta(x)$ and $\hat{f}_\theta(x,t)$. This requires some sense of uniqueness in this separation, so certain restrictions need be put on the model to consider systems less general than \eqref{eq:noncangen}. A major advantage that comes with this feature of the PHNN approach is that it makes it possible to learn a model of the system as if under ideal conditions even if data is sampled from a system affected by disturbances, if one assumes that an undisturbed system is closed and thus given only by the internal dynamics.

The motivating idea behind the present paper is to extend the framework of \cite{Eidnes2023pseudo} to PDEs. In principle, one could always treat the spatially discretized PDE as a system of ODEs and apply the PHNN models of \cite{Eidnes2023pseudo} to that. However, that would be disregarding certain structures we know to be present in the discretized PDE and would lead to inefficient models. Thus, compared to the ODE case, we consider different neural network architectures. Moreover, we will in the PDE setting impose some restrictions on the form of the external forces, in that we will not allow for them to depend on spatial derivatives of the solution. On the other hand, we will consider a more general form of the internal dynamics, where dissipation can result from a separate term and not just damping of the Hamiltonian. This mean that we can model metriplectic systems, in addition to Hamiltonian, port-Hamiltonian and dissipative systems.

Although HNN and extensions of this have attracted considerable attention in recent years, there has been very few studies on extending the methodology to PDEs. To our knowledge, the only prior works that consider HNN for PDEs are those of Matsubara et al.\ in \cite{Matsubara2020deep} and Jin et al.\ in \cite{Jin2022learning}. The latter has included a numerical example on the nonlinear Schrödinger equation. The former reference considers both Hamiltonian PDEs, exemplified by the Korteweg--de Vries equation, and an extension to dissipative PDEs, demonstrated on the Cahn--Hilliard equation. That paper has been a major inspiration for our work, especially on the neural network architecture we use to model the integrals in our PDE formulation. By generalizing to a wider class of PDEs that can have conservative and dissipative terms at once, and also allowing for external forces, we largely expand the utility of this learning approach.

The extension of PHNN to PDEs we propose here should naturally also be put in context with other recent advances in learning of PDEs from data. Long et al.\ introduced PDE-Net in \cite{Long2018pde}, and together with \cite{Matsubara2020deep} this may be the work that is most comparable to what we present here. Their model is similar to the baseline model we will compare PHNN to in this paper, albeit less general. Their approach has two components: learning neural network models for the nonlinear terms in the PDE and identifying convolution operators that correspond to appropriate finite difference approximations of the spatial derivative operators present. They do however make considerable simplifying assumptions in their numerical experiments, e.g.\ only considering linear terms and a forward Euler discretization in time. Other works that have received significant attention are those that have focused on identifying coefficients of terms of the PDE, both in the setting where one assumes that the terms are known and approximate them by neural network models \cite{Raissi2019physics} and in the setting where one also identifies the terms present from a search space of candidates using sparse regression \cite{Rudy2017data, Schaeffer2017learning, Kaheman2020sindy}. There has also been considerable recent research on learning operators associated with the underlying PDEs, where two prominent methods are Fourier neural operators (FNO) \cite{Anandkumar2020neural, Li2021fourier} and deep operator networks (DeepONet) \cite{Lu2021learning}. These operators can e.g.\ map from a source term to solution states or from an initial state to future states; in the latter case, learning the operator equates to solving the forward problem of the PDE. The review paper \cite{Brunton2023machine} summarizes the literature on operator learning and system identification of PDEs, as well as recent developments on learning order reductions.

As will be demonstrated theoretically and experimentally in this paper, assuming a pseudo-Hamiltonian structure when solving the inverse problem for PDEs has both qualitative and quantitative advantages. The latter is shown by numerical comparisons to a baseline model on five test cases. The main qualitative feature of PHNN is that it is composed of up to six trainable submodels, which after training each can be studied for an increased understanding of the system we are modelling. And if initial experiments for instance indicate that the system is Hamiltonian, we can retrain with this assumption imposed and thus learn more accurate solutions by pure HNN models. Moreover, we could train a system affected by external forces and remove these from the model after training, so that we have a model unaffected by these disturbances.
The code for this paper is built on the code for PHNN for ODEs developed for \cite{Eidnes2023pseudo}, and we have updated the GitHub repository \url{https://github.com/SINTEF/pseudo-hamiltonian-neural-networks} and the Python package \texttt{phlearn} with this extension to the PDE case.

The rest of this paper is organised as follows. In the next section, we explore the theoretical foundations upon which our method is based. Then the pseudo-Hamiltonian formulation and the class of PDEs we will learn are presented and discussed in Section \ref{sec:phform}. Section \ref{sec:phnn} is the centerpiece of the paper, as it is here we present the PHNN method for PDEs. We then dedicate a substantial portion of the paper to presenting and evaluating numerical results for various PDEs, in Section \ref{sec:experiments}. The penultimate section is devoted to analysis of the results and our model, and a discussion of open questions to address in future research. We summarize the main results and draw some conclusions in the last section.
\section{Background: Derivatives, discretizations and neural networks}\label{sec:background}
Before delving into the pseudo-Hamiltonian formulation and the model we propose based on this, we will review and discuss some requisites for making efficient neural network models of systems governed by PDEs.

\subsection{Learning dynamical systems}
Consider the first-order in time and $p$-order in space PDE
\begin{equation}\label{eq:pde}
    u_t = g(u^\alpha, x, t), \quad  u \in H^p(\Omega), x \in \Omega \subseteq \mathbb{R}^d, t \in \mathbb{R},
\end{equation}
with
\begin{equation*}
    u^\alpha = \left\{ \frac{\partial^{|\alpha|} u}{\partial x_1^{\alpha_1} x_2^{\alpha_2} \cdots x_d^{\alpha_d}} : |\alpha| \leq p \right\}, \quad \alpha \in (\mathbb{Z}^{\geq})^d.
\end{equation*}
We seek to train a model $\hat{g}_\theta$ of $g$ so that solving $u_t = \hat{g}_\theta(u^\alpha, x, t)$ leads to accurate predictions of the future states of the system. The universal approximation theorem \cite{Hornik1989multilayer, Cybenko1989approximation} states that $g$ can be approximated with an arbitrarily small error by a neural network. In practice, we have to assume an abundance of observations of $u_t$ and $u^\alpha$ at $t$ and $x$ to actually find a precise neural network approximation of $g$. This brings us straight to one of the fundamental challenges of machine learning of differential equations: in a typical real-world setting, we cannot expect to have data on the derivatives, neither temporal nor spatial. Thus we will have to depend on approximations obtained from discrete data. In this paper we will use sub- and superscript to denote discrete solution points in space resp.\ time. That is, $u^\alpha(x) = u(x,t^j)$ and $u_i(t)=u(x_i,t)$, and we will suppress the arguments when they are not necessary. Let us consider the issue of time-discretization first, an issue shared by ODEs and PDEs alike, and defer the second issue to the next subsection.

In several of the papers introducing the most prominent recent methods for learning finite-dimensional dynamical systems, e.g.\ the original HNN paper \cite{Greydanus2019hamiltonian} and the first paper by Brunton et al.\ on system identification \cite{Brunton2023machine}, the derivatives of the solution are assumed to be known or approximated by finite differences. Approximating the time-derivative by the forward finite difference operator is equivalent to training using the forward Euler integrator, which is also what is done in the PDE-Net papers \cite{Long2018pde, Long2019pde}. However, there has been several recent papers proposing more efficient training methods that incorporate other numerical integration schemes, see e.g. \cite{Chen2019symplectic, Jin2020sympnets, David2021symplectic}. We follow \cite{Eidnes2023pseudo, Noren2023learning} and set up the training is such a way that we can use any mono-implicit integrator; that is, any integrator that relies explicitly on the solution in the times it integrates from and to. For the majority of the experiments in this paper, we use the implicit midpoint method, which is second-order, symplectic and symmetric. That is, we train the model $\hat{g}_\theta$ by identifying the parameters $\theta$ that minimize the loss function
\begin{equation*}
    \mathcal{L}_{g_\theta} = \bigg\lVert \frac{u^{j+1}-u^{j}}{\Delta t} - \hat{g}_\theta\Big(\frac{(u^\alpha)^{j}+(u^\alpha)^{j+1}}{2}, x,\frac{t^j + t^{j+1}}{2}\Big) \bigg\rVert_2^2,
\end{equation*}
given for one training point $u^\alpha$ and barring regularization for now.  This yields a considerable improvement over the forward Euler method at next to no additional computational cost, since the model in both cases is evaluated at only one point at each iteration of training. The option to use other integrators, including symmetric methods of order four and six, is readily implemented in the \texttt{phlearn} package, and we do demonstrate the need for and utility of a fourth-order integrator in Section \ref{sec:pm}. For a thorough study of integrators especially suited for training neural network models of dynamical systems, we refer the reader to \cite{Noren2023learning, Noren2023learning2}.

\subsection{Spatial derivatives and convolution operators}
Moving from finite-dimensional systems to infinite-dimensional systems introduces the issue of how to approximate spatial derivatives by the neural network models. Thankfully, a proposed solution to this issue can be found in recent literature, as several works have noted the connection between finite difference schemes for differential equations and the convolutional neural network models originally developed for image analysis; see \cite{Cai2012image, Dong2017image, Ruthotto2020deep, Bar2019learning, Celledoni2023predictions}.

Given a function $u$ and a kernel or filter $w$, a discrete convolution is defined by
\begin{equation}
(u * w)(x_i) = \sum_{j=-r}^{s} w_j  u(x_{i-j}), \quad r, s \geq 0.
\end{equation}
Here $*$ is called the convolution operator, and the kernel $w$ is a tensor containing trainable weights: $w = \left[w_{-r}, w_{-r+1}, \ldots, w_0, \ldots, w_{s-1}, w_{s}\right]$. If the function $u$ is periodic, so that $u(x_i) = u(x_{i+M})$ for some $M$, we obtain a circular convolution, which can be expressed by a circulant matrix applied on the vector $u = \left[u_0, \ldots, u_{M-1}\right]^T$, where $u_i:= u(x_i)$.

A convolutional layer in a neural network can be represented as
\begin{equation*}
y_k(u_i) = \phi\big( (u * w_{k})(x_i) + b_k\big) = \phi\bigg(\sum_{j=-r}^{s} w_{kj} u_{i-j} + b_k\bigg),
\end{equation*}
where $y_k(u_i)$ is the output of the $k$-th feature map at point $u_i$, $w_{kj}$ are the weights of the kernel $w_k$, $b_k$ is the bias term, and $\phi(\cdot)$ is an activation function. The width of the layer is $r+s+1$, and this is usually referred to as the size of the convolution kernel, or filter. For our purpose it makes sense to have either $r=0$ or $r=s$, and the latter is the standard when convolutional neural networks are used in image analysis. Training the convolutional layer of a neural network constitutes of optimizing the weights and biases, which we collectively denoted by $\theta$ in the previous subsection.

Similarly, a finite difference approximation of the $n$-th order derivative of $u$ at a point $x_i$ can also be expressed as applying a discrete convolution:
\begin{equation}
\frac{\de^n u(x_i)}{\de x^n} \approx \sum_{j=-r}^{s} a_j u(x_{i-j}),
\end{equation}
where the finite difference weights $a_j$ depend on the spatial grid. If we assume the spatial points to be equidistributed and let $h:=x_{i+1}-x_i$, we have e.g.
\begin{align*}
\frac{\de u(x_i)}{\de x} &= \frac{u(x_{i+1})-u(x_{i})}{h} + \mathcal{O}(h),\\
\frac{\de u(x_i)}{\de x} &= \frac{u(x_{i+1})-u(x_{i-1})}{2h} + \mathcal{O}(h^2),\\
\frac{\de^2 u(x_i)}{\de x^2} &= \frac{u(x_{i+1})-u(x_i)+u(x_{i-1})}{h^2} + \mathcal{O}(h^2),\\
\frac{\de^3u(x_i)}{\de x^3} &= \frac{u(x_{i+2}) - 2u(x_{i+1}) + 2u(x_{i-1}) - u(x_{i-2})}{2h^3} + \mathcal{O}(h^2).
\end{align*}
Hence, a kernel size of two, with e.g.\ $r=0$ and $s=1$, is sufficient to obtain a first-order approximation of the first derivative, while a kernel size of three is sufficient and necessary to obtain second order approximations of first and second derivatives. Further, kernel size five is needed to approximate the third derivative. As noted by \cite{Ruthotto2020deep}, higher-order derivatives can be approximated either by increasing the kernel size or applying multiple convolution operations. In our models, we have designed neural networks where only the first layer is convolutional, and thus the kernel size restricts the order of the derivative we can expect to learn, while it also restricts the order of the approximations of these derivatives.

\subsection{Variational derivative}
Given the function $H$ depending on $u$, $x$ and the first derivative $u_x$, let $\mathcal{H}$ be the integral of  $H$ over the spatial domain:
\begin{equation}\label{eq:integral}
\mathcal{H}[u] = \int_\Omega H(x,u,u_x) \, \de x.
\end{equation}
The variational derivative, or functional derivative, $\frac{\delta \mathcal{H}}{\delta u}[u]$ of $\mathcal{H}$ is defined by the property
\begin{equation}\label{eq:varder}
\left\langle \frac{\delta \mathcal{H}}{\delta u}[u], v\right\rangle_{L_2} = \frac{\de}{\de\epsilon}\bigg\rvert_{\epsilon=0} \mathcal{H}[u+\epsilon v] \quad \forall v \, \in H^p(\Omega).
\end{equation}
When $\mathcal{H}$ as here only depends on first derivatives, the variational derivative can be calculated explicitly by the relation
\begin{equation*}
\frac{\delta \mathcal{H}}{\delta u}[u] = \frac{\partial H}{\partial u} - \frac{\de}{\de x} \frac{\partial H}{\partial u_x},
\end{equation*}
assuming enough regularity in $H$.
\section{Pseudo-Hamiltonian formulation of PDEs}\label{sec:phform}

In this paper we consider the class of PDEs that can be written on the form
\begin{equation}\label{eq:pdeclass}
u_t = S(u^\alpha, x)\dfrac{\delta \mathcal{H}}{\delta u}[u] - R(u^\alpha, x) \dfrac{\delta \mathcal{V}}{\delta u}[u] + f(u^\alpha,x,t),
\end{equation}
where $S(u^\alpha,x)$ and $R(u^\alpha,x)$ are operators that are skew-symmetric resp.\ positive semi-definite with respect to the $L^2$ inner product, $\mathcal{H}$ and $\mathcal{V}$ are integrals of the form \eqref{eq:integral} and $f : \mathbb{R} \times \mathbb{R}^d \times \mathbb{R} \rightarrow \mathbb{R}$.
To be consistent with our previous work \cite{Eidnes2023pseudo} and to make clear the connection to the vast recent literature on Hamiltonian neural networks, we say that \eqref{eq:pdeclass} is the class of \textit{pseudo-Hamiltonian PDEs}. This marks a generalization of the definition used in \cite{Eidnes2023pseudo}, in addition to the extension to infinite-dimensional systems, in that we here allow for $\mathcal{H}$ and $\mathcal{V}$ to be two different integrals.

The naming of this class is a challenge, since similar but not identical classes have been called by a myriad of names in the literature. Ignoring the term $f$, the class could be referred to as metriplectic PDEs, where the name is a portmanteau of metric and symplectic \cite{Guha2007metriplectic, Bloch2013gradient}. Examples of metriplectic PDEs are the viscous Burgers' equation and the Navier--Stokes equation. The formulation is also similar to an infinite-dimensional variant of the General Equation for Non-Equilibrium Reversible-Irreversible Coupling (GENERIC) formalism from thermodynamics \cite{Grmela1997dynamics, Ottinger1997dynamics}, except for $f$ and the fact that $R(u^\alpha,x)$ is positive instead of negative semi-definite. Furthermore, the GENERIC formalism requires the degeneracy conditions
\begin{equation*}
    R(u^\alpha, x)\dfrac{\delta \mathcal{H}}{\delta u}[u] = S(u^\alpha, x) \dfrac{\delta \mathcal{V}}{\delta u}[u] = 0
\end{equation*}
to be satisfied. We do not assume this to be satisfied and thus do not impose this condition on our model, but we consider that a highly relevant future extension of our work. In the finite-dimensional case, neural networks that preserve the GENERIC formalism have been studied in \cite{Zhang2022gfinns}.

 In the case $\mathcal{V} =0$ and $f(u^\alpha,x,t) = 0$, we have the class of integral-preserving PDEs, which encompasses all (non-canonical) Hamiltonian PDEs \cite{Leimkuhler2004simulating}. That is, given the appropriate boundary conditions, e.g.\ periodic, the PDE will preserve the integral $\mathcal{H}$, usually labeled the integral of motion, of the system. This follows from the skew-symmetry of $S$:
\begin{align*}
\dfrac{\de\mathcal{H}}{\de t} = \left\langle \dfrac{\delta \mathcal{H}}{\delta u}[u], \dfrac{\partial u}{\partial t}\right\rangle_{L^2} = \left\langle \dfrac{\delta \mathcal{H}}{\delta u}[u], S(u^\alpha,x)\dfrac{\delta \mathcal{H}}{\delta u}[u]\right\rangle_{L^2}= 0.
\end{align*}
If $S$ in addition satisfies the Jacobi identity and thus defines a Poisson bracket, $\mathcal{H}$ is a Hamiltonian of the system \cite{Olver1993applications}. If $\mathcal{H}=0$ and $f(u^\alpha,x,t) = 0$ but $\mathcal{V} \geq 0$, the PDE \eqref{eq:pdeclass} will dissipate the integral $\mathcal{V}$, and $\mathcal{V}$ may be called a Lyapunov function.

The general pseudo-Hamiltonian formulation does not in itself have a geometric structure. The motivation for still considering this formulation is two-fold: i) to develop a general machine learning model where geometric structures can be imposed to handle different system classes, including Hamiltonian, port-Hamiltonian, dissipative and metriplectic PDEs, with and without external forces; ii) to obtain grey-box models with parts that can be studied separately to understand more about the system.

\begin{example}\label{ex:kdvburgers}
Consider the KdV--Burgers (or viscous KdV) equation \cite{Wang1996exact}
\begin{equation}\label{eq:kdvburgers}
u_t + \eta u u_x - \nu u_{xx} - \gamma^2 u_{xxx} = 0.
\end{equation}
This is a metriplectic PDE that can be written on the form \eqref{eq:pdeclassspef} with $A$ and $R$ both being the identity operator $I$, $S = \frac{\partial}{\partial x}$ and  $f(u,x,t)=0$, and
\begin{equation}\label{eq:kdvburgersH}
\mathcal{H} = - \int_\Omega \left(\frac{\eta}{6}u^3 + \frac{\gamma^2}{2}u_x^2\right) \, dx
\end{equation}
and
\begin{equation}\label{eq:kdvburgersV}
\mathcal{V} = \frac{\nu}{2}\int_\Omega u_x^2 \, dx.
\end{equation}
We see this connection by deriving the variational derivatives
\begin{equation}\label{eq:kdvburgers_varderH}
\dfrac{\delta \mathcal{H}}{\delta u}[u] = - (\frac{\eta}{2} u^2  - \gamma^2 u_{xx})
\end{equation}
and
\begin{equation}\label{eq:kdvburgers_varderV}
\dfrac{\delta \mathcal{V}}{\delta u}[u] = - \nu u_{xx}.
\end{equation}
We have that \eqref{eq:kdvburgers} reduces to the inviscid Burgers' equation for $\eta = -1$ and $\nu=\gamma=0$,  the viscous Burgers' equation for $\eta = -1$, $\nu\neq 0$ and $\gamma=0$, and the Korteweg--de Vries (KdV) equation for $\nu=0$, $\eta \neq 0$ and $\gamma \neq 0$.
\end{example}

\subsection{Spatial discretization}
In this section and this section only we will use boldface notation for vectors, to distinguish continuous functions and parameters from their spatial discretizations. Assume that values of $u$ are obtained at grid points $\mathbf{x} = [x_0, \ldots, x_M]^T$. Following \cite{Eidnes2018adaptive}, we interpret these as quadrature points with non-zero quadrature weights $\mathbf{\kappa} = [\kappa_0, \ldots, \kappa_M]^T$, and approximate the $L_2$ inner product by a weighted discrete inner product:
\begin{equation*}
\langle u,v\rangle = \int_\Omega u(x) v(x) \, \mathrm{d}x \approx \sum_{i=0}^M \kappa_i u(x_i)v(x_i) = u^T \text{diag}(\kappa) v =: \langle u,v \rangle_\kappa.
\end{equation*}
Let $\mathbf{p}$ denote the discretization parameters that consist of $\mathbf{x}$ and the associated $\kappa$. Then, assuming that there exists a consistent approximation $\mathcal{H}_\mathbf{p}(\mathbf{u})$ to $\mathcal{H}[u]$ that depends on $u$ evaluated at $\mathbf{x}$, we define the discretized variational derivative by the analogue to \eqref{eq:varder}
\begin{equation*}
\left\langle \frac{\delta \mathcal{H}_\mathbf{p}}{\delta \mathbf{u}}(\mathbf{u}), \mathbf{v}\right\rangle_{\kappa} = \frac{\de}{\de\epsilon}\bigg\rvert_{\epsilon=0} \mathcal{H}_\mathbf{p}(\mathbf{u}+\epsilon \mathbf{v}) \quad \forall \mathbf{v} \, \in \mathbb{R}^{M+1}.
\end{equation*}
Thus, as shown in \cite{Eidnes2018adaptive}, we have a relationship between the discretized variational derivative and the gradient:
\begin{equation*}
\frac{\delta \mathcal{H}_\mathbf{p}}{\delta \mathbf{u}}(\mathbf{u}) = \text{diag}(\kappa)^{-1} \nabla_{\mathbf{u}} \mathcal{H}_\mathbf{p}(\mathbf{u}).
\end{equation*}
Furthermore, we approximate $S(u^\alpha,x)$ and $R(u^\alpha,x)$ by matrices $S_d(\mathbf{u})$ and $R_d(\mathbf{u})$ that are skew-symmetric resp.\ positive semi-definite with respect to $\langle\cdot,\cdot\rangle_\kappa$. Then a spatial discretization of \eqref{eq:pdeclass} is given by
\begin{equation*}
\mathbf{u}_t = S_d(\mathbf{u})\dfrac{\delta \mathcal{H}_\mathbf{p}}{\delta \mathbf{u}}(\mathbf{u})- R_d(\mathbf{u}) \dfrac{\delta \mathcal{V}_\mathbf{p}}{\delta \mathbf{u}}(\mathbf{u}) + \mathbf{f}(\mathbf{u},\mathbf{x},t),
\end{equation*}
which may equivalently be written as
\begin{equation}\label{eq:odesys}
\mathbf{u}_t = S_\mathbf{p}(\mathbf{u})\nabla_\mathbf{u}\mathcal{H}_\mathbf{p}(\mathbf{u}) - R_\mathbf{p}(\mathbf{u}) \nabla_\mathbf{u}\mathcal{V}_\mathbf{p}(\mathbf{u}) + \mathbf{f}(\mathbf{u},\mathbf{x},t),
\end{equation}
where $S_\mathbf{p}(\mathbf{u}) := S_d(\mathbf{u})\, \text{diag}(\kappa)^{-1}$ and $R_\mathbf{p}(\mathbf{u}) := R_d(\mathbf{u})\, \text{diag}(\kappa)^{-1}$ are skew-symmetric resp.\ positive semi-definite by the standard definitions for matrices.

Thus, upon discretizing in space, we obtain a system of ODEs \eqref{eq:odesys} that is on a form quite similar to the generalized pseudo-Hamiltonian formulation considered in \cite{Eidnes2023pseudo}. In fact, if $\mathcal{V} = \mathcal{H}$, we obtain the system
\begin{equation*}
\mathbf{u}_t = (S_\mathbf{p}(\mathbf{u}) - R_\mathbf{p}(\mathbf{u})) \nabla_\mathbf{u}\mathcal{H}_\mathbf{p}(\mathbf{u}) + \mathbf{f}(\mathbf{u},\mathbf{x},t).
\end{equation*}
Still, we do not recommend applying the PHNN method of \cite{Eidnes2023pseudo} on this directly without taking into consideration what we know about $\mathcal{H}_\mathbf{p}$. Specifically, we want to exploit that it is a discrete approximation of the integral \eqref{eq:integral}, and can thus be expected to be given by a sum of $M$ terms that each depend in the same way on $u_i$ and the neighbouring points $u_{i-1}$ and $u_{i+1}$. Hence, as discussed in the next section, we will employ convolutional neural networks with weight sharing across the spatial discretization points.

\begin{example}\label{ex:kdvburgersdisc}
Consider again the KdV--Burgers equation \eqref{eq:kdvburgers}, on the domain $\Omega = [0,P]$ with periodic boundary conditions $u(0,t) = u(P,t)$. We assume that the $M+1$ grid points are equidistributed and define $h:=x_{i+1}-x_i = P/M$. We approximate the integrals \eqref{eq:kdvburgersH} and \eqref{eq:kdvburgersV} by
\begin{align}\label{eq:kdvH}
\mathcal{H}_\mathbf{p} = - h \sum_{i=0}^{M-1} \left(\frac{\eta}{6}u_i^3 + \frac{\gamma}{2}\left(\delta_f u_i\right)^2\right)
\end{align}
and
\begin{align}\label{eq:kdvV}
\mathcal{V}_\mathbf{p} = \frac{\nu}{2} h \sum_{i=0}^{M-1} \kappa_i \left(\delta_f u_i\right)^2,
\end{align}
where the operator $\delta_f$ denotes forward difference, i.e. $\delta_f u_i = (u_{i+1} - u_i)/h$. Furthermore, we approximate $\partial_x$ by the matrix corresponding to the central difference approximation $\delta_c$ defined by $\delta_c u_i = (u_{i+1}-u_{i-1})/(2 h)$, i.e.
\begin{equation*}
S_d = 
\frac{1}{2 h}
\begin{pmatrix}
    0 & 1 & 0 & \cdots & 0 &  -1\\
    -1 & 0 & 1 & 0 & \cdots&\\
    0 & -1 & 0 & 1 & 0 & \cdots\\
    & & \ddots & \ddots & \ddots &\\
    & \cdots & 0 & -1 & 0 & 1 \\
    1 & 0 & \cdots & 0 & -1 & 0
\end{pmatrix} \in \mathbb{R}^{M \times M},
\end{equation*}
where the first and last rows are adjusted according to the periodic boundary conditions.

To obtain \eqref{eq:odesys}, we have that $S_\mathbf{p} = \frac{1}{h} S_d$ and $R_\mathbf{p} = \frac{1}{h} I$, with $I$ being the identity matrix, and take the gradients of the approximated integrals to find
\begin{align*}
\nabla_u \mathcal{H}_\mathbf{p} &= -h \left( \frac{\eta}{2} \mathbf{u}^2 - \gamma^2 \delta_c^2 \mathbf{u} \right),\\
\nabla_u \mathcal{V}_\mathbf{p} &= -h \, \nu \, \delta_c^2 \mathbf{u},
\end{align*}
where $\mathbf{u}^2$ and $\mathbf{u}^3$ denote the element-wise square and cube of $\mathbf{u}$, and $\delta_c^2 := \delta_f\delta_b$ denotes the second-order difference operator approximating the second derivative by $\delta_c^2 u_i = (u_{i+1}- 2u_i + u_{i-1})/(2 h)$.
Observe that $\frac{\delta \mathcal{H}_\mathbf{p}}{\delta \mathbf{u}}(\mathbf{u}) = \frac{1}{h} \nabla_{\mathbf{u}} \mathcal{H}_\mathbf{p}(\mathbf{u})$ and $\frac{\delta \mathcal{V}_\mathbf{p}}{\delta \mathbf{u}}(\mathbf{u}) = \frac{1}{h} \nabla_{\mathbf{u}} \mathcal{V}_\mathbf{p}(\mathbf{u})$ are consistent discrete approximations of \eqref{eq:kdvburgers_varderH} and \eqref{eq:kdvburgers_varderV}. Moreover, they are second-order approximations of these variational derivatives, even though \eqref{eq:kdvH} and \eqref{eq:kdvV} are only first-order approximations of the integrals \eqref{eq:kdvburgersH} and \eqref{eq:kdvburgersV}.
\end{example}

\subsection{Restricting the class by imposing assumptions}\label{sec:restrict}
Without imposing any further restrictions, the formulation \eqref{eq:pdeclass} can be applied to any PDE that is first-order in time and will not be unique for any system. Any contribution from the two first terms on the left-hand side could also be expressed in $f$, and even if we restrict this term, the operators $S$ and $R$ are generally not uniquely defined for a corresponding integral. In the remainder of this paper, we will consider the cases where the operators $S$ and $R$ are linear and independent of $x$ and $u$, we assume $R$ to be symmetric, and we will not let $f$ depend on derivatives of $u$. Furthermore, we apply the symmetric positive semi-definite operator $A$ to the equation and require that this commutes with $R$ and $S$. We redefine $S := A S$, $R:=AR$ and $f(u,x,t) := A f(u,x,t)$, and thus get
\begin{equation}\label{eq:pdeclassspef}
A u_t = S \dfrac{\delta \mathcal{H}}{\delta u}[u] - R \dfrac{\delta \mathcal{V}}{\delta u}[u] + f(u,x,t),
\end{equation}
where the new $S$ is still skew-symmetric and the new $R$ is still symmetric and positive semi-definite.

In the following we will denote the identity operator by $I$ and the zero operator by $0$, so that $I v = v$ and $0 v = 0$ for any $v \in L_2$. We note that the zero operator is positive semi-definite, symmetric and skew-symmetric, while the identity operator is symmetric and positive semi-definite, but not skew-symmetric.
\section{The PHNN model for PDEs}\label{sec:phnn}
Since we assume that the operators $A$, $S$ and $R$ are independent of $x$ and $u$, the discretization of these operators will necessarily result in circulant matrices, given that $u$ is periodic. That is, they can be viewed as discrete convolution operators. We thus set $\hat{A}_\theta^{[k_1]}$, $\hat{S}_\theta^{[k_2]}$ and $\hat{R}_\theta^{[k_3]}$ to be trainable convolution operators, where $k_1$, $k_2$ and $k_3$ denote the kernel sizes, and we impose symmetry on $\hat{A}_\theta^{[k_1]}$ and $\hat{R}_\theta^{[k_3]}$ and skew-symmetry on $\hat{S}_\theta^{[k_2]}$.
Furthermore, we let $\hat{\mathcal{H}}_\theta$ and $\hat{\mathcal{V}}_\theta$ be two separate neural networks that take input vectors of length $M$, the number of spatial discretization points, and output a scalar. The neural network $\hat{f}_\theta$ can take input vectors representing both $u$, $x$ and $t$ and outputs a vector of length $M$.

The full pseudo-Hamiltonian neural network model for PDEs is then given by
\begin{equation}\label{eq:phnn}
    \hat{g}_\theta(u,x,t) = (\hat{A}_\theta^{[k_1]})^{-1} \Big(\hat{S}_\theta^{[k_2]}\nabla\hat{\mathcal{H}}_\theta(u) - \hat{R}_\theta^{[k_3]} \nabla\hat{\mathcal{V}}_\theta(u) + k_4 \hat{f}_\theta(u,x,t)\Big),
\end{equation}
where we also have introduced $k_4$, which should be $1$ or $0$ depending on whether or not we want to learn a force term. Given a set of $N$ training points $\{(u^{j_n}, u^{j_n+1}, t^{j_n})\}_{n=1}^N$ varying across time and different stochastic realizations of initial conditions, we let the loss function be defined as 
\begin{equation}
\label{eq:loss_ode}
    \mathcal{L}_{g_\theta}(\{(u^{j_n}, u^{j_n+1}, t^{j_n})\}_{n=1}^N) = \frac{1}{N}\sum_{n=1}^N\bigg| \frac{u^{j_n+1}-u^{j_n}}{\Delta t} - \hat{g}_\theta\Big(\frac{u^{j_n}+u^{j_n+1}}{2}, x,\frac{t^{j_n} + t^{j_n+1}}{2}\Big) \bigg|^2,
\end{equation}
if the implicit midpoint integrator is used. For the experiments in Section \ref{sec:pm} we use the fourth-order symmetric integrator introduced in \cite{Eidnes2023pseudo}, and the loss function is amended accordingly.

\subsection{Implementation}

PHNN is comprised of up to six trainable models; the framework is very flexible and assumptions may be imposed so one or more of the parts do not have to be learned. Moreover, careful considerations should be made on how to best model the different parts. In the following, we explain how we have set up the models in our code.

\subsubsection{Modelling $\mathcal{H}$ and $\mathcal{V}$}
The networks $\hat{\mathcal{H}}_\theta$ and $\hat{\mathcal{V}}_\theta$ take inputs of dimension $M$, the number of spatial discretization points, and consist of one convolutional layer with kernel size two followed by linear layers corresponding to convolutional layers with kernel size one, and then in the last layer performs a summation of the $M$ inputs to one scalar. Following each of the first two layers, we apply the $\tanh$ activation function.
To impose the periodic boundary conditions, we pad the input to the convolutional layer by adding $u(P) = u(0)$ at the end of the array of the discretized $u$. A similar technique was suggested in \cite{Matsubara2020deep}, although they use a kernel of size three on the first convolutional layer. We opt to have a smaller filter, since kernel size two is sufficient to learn the forward difference approximation of the first derivative in the integrals, which in turn is sufficient to obtain second order approximation of the resulting variational derivative; this is shown for the KdV--Burgers equation in Example \ref{ex:kdvburgersdisc}. If we want to be able to learn derivatives of order two in the integral, we would need kernel size three, and to pad the input with one element on each side. If we want to learn derivatives of order three or four, or if we want to learn third- or fourth-order approximations of the derivatives, we would need the kernel size of the convolutional layer to be five, and to pad the input by two elements on each side. This adjustment can easily be made in our code. For the examples in this paper, we would not gain anything by increasing the kernel size, because we only have up to first derivatives in the integrals and because the training data is generated using second order spatial discretizations. On the other hand, having a kernel of size two simplifies the learning and may facilitate superior performance over a model that does not rely on a pseudo-Hamiltonian structure and have to approximate up to third derivatives by convolutional neural networks.

\subsubsection{Modelling $A$, $S$, $R$ and $f$}
\label{sububsec:modelling_kernel_size}
In \eqref{eq:phnn}, $k = (k_1,k_2,k_3,k_4)$ are hyperparameters that determine the expressiveness of the model. Setting $k_1=k_2=k_3=M$ and $k_4=1$ means that we can approximate the general system \eqref{eq:pdeclassspef}, while setting $k_1=k_3=1$, $k_2=3$ and $k_4=0$ would be sufficient to learn a model for the discretized KdV--Burgers system \eqref{eq:kdvburgers}. In fact, if we set $k_3=3$, the skew-symmetric operator $\hat{S}_\theta^{[k_2]}$ is uniquely defined up to a multiplicative constant, since we require $w_0 = 0$ and $w_1=-w_{-1}$. Moreover, since this constant would only amount to a scaling between the operator and the discrete variational derivative it is applied to, $\hat{S}_\theta^{[k_2]}$ does not have to be trained in this case; determining $w_1$ would just lead to a scaling of the second-order approximation of the first derivative in space that could be compensated by a scaling of $\mathcal{H}$. Similarly, if the kernel size of $A$ or $R$ is $3$, we could set $w_0=1$ and learn a single parameter $w_1 = w_{-1}$ for each of these when training the model. This corresponds to learning a linear combination of the identity and the second-order approximation of the second derivative in space.

We model $f$ by the neural network $\hat{f}_\theta$ that may take either of the variables $u$, $x$ and $t$ as input. This has three linear layers, i.e.\ convolutional layers with kernel size one, with the $\tanh$ activation function after each of the first two. If $\hat{f}_\theta$ depends on $x$, periodicity on the domain $[0,P]$ is imposed in a similar fashion as suggested in \cite{Zhang2020learning, Lu2021physics, Lu2022comprehensive} for hard-constraining periodic boundary conditions in physics-informed neural networks and DeepONet. That is, we replace the input $x$ by the first two Fourier basis functions, $\sin{\big(\frac{2\pi}{P}x\big)}$ and $\cos{\big(\frac{2\pi}{P}x\big)}$, which is sufficient for expressing any $x$-dependent periodic function.

In the numerical experiments of the next section, we do not consider systems where $A$ and $R$ are anything other than linear combinations of the identity and the spatial second derivative, or $S$ is anything other than the first derivative in space. Thus we set $k=[3,3,3,1]$ in our most general model, which is expressive enough to learn all the systems we consider. We also consider what we call \textit{informed} PHNN models where we assume to have prior knowledge of the operators, affecting $k$, and also what variables $f$ depend on.

\subsubsection{Leakage of constant}\label{sec:leakage}
If $R$ is the identity, or a linear combination of the identity with differential operators, the separation between the dissipation term and the external force term in \eqref{eq:pdeclassspef} is at best unique up to a constant, which means that there may be leakage of a constant between the two last terms of the PHNN model. Hence, we must make some assumptions about these terms to separate them as desired. If we want the external force term to be small, we may use regularization and penalize large values of $\lVert\hat{f}_\theta\rVert$ during training. The option to do this is implemented in the \texttt{phlearn} package. However, for the numerical experiments in the next section, we have instead opted to assume that the dissipative term should be zero for the zero solution, and thus correct the two terms in question after training so that it adheres to this without changing the full model. That is, if we have the model
\begin{equation}\label{eq:PHNNpre}
    \hat{g}^\text{pre}_\theta(u,x,t) = (\hat{A}_\theta^{[k_1]})^{-1} \Big(\hat{S}_\theta^{[k_2]}\nabla\hat{\mathcal{H}}_\theta(u) - \hat{R}_\theta^{[k_3]} \nabla\hat{\mathcal{V}}^\text{pre}_\theta(u) + k_4 \hat{f}^\text{pre}_\theta(u,x,t)\Big),
\end{equation}
when the last training step is performed, we set
\begin{align*}
    \nabla\hat{\mathcal{V}}_\theta(u) &:= \nabla\hat{\mathcal{V}}^\text{pre}_\theta(u) - k_4 \nabla\hat{\mathcal{V}}^\text{pre}_\theta(0),\\
    \hat{f}_\theta(u,x,t) &:= \hat{f}^\text{pre}_\theta(u,x,t) - \hat{R}_\theta^{[k_3]}\nabla\hat{\mathcal{V}}^\text{pre}_\theta(0)
\end{align*}
to get our final model \eqref{eq:phnn}, which is equivalent to \eqref{eq:PHNNpre}. Then we may remove the dissipation or external forces from the model simply by setting $k_3=0$ or $k_4=0$. Note, however, that this correction may not work as expected if the zero solution is far outside the domain of the training data, since the neural network $\hat{\mathcal{V}}^\text{pre}_\theta$ like most neural networks generally extrapolates poorly. In that case, regularization is to be preferred.

\subsubsection{Algorithms}
We refer to Algorithm \ref{alg:phnn_train} and Algorithm \ref{alg:baseline_train} for the training of the PHNN and baseline models, respectively.

\SetKwComment{Comment}{/* }{ */}
\begin{algorithm}
\caption{The training phase of the PHNN algorithm}\label{alg:phnn_train}
\KwData{Observations $D=\{(t_1, \vec{x}^1, \vec{u}^1), \ldots, (t_N, \vec{x}^N, \vec{u}^N\}$}
\KwData{Number of epochs $K$}
\KwData{Batch size $M_b$}
\KwData{Initial CNN $\hat{H}_\theta, \hat{V}_\theta$}
\KwData{Initial DNN $\hat{f}_\theta$}
\KwData{Matrices $\hat{A}_\theta^{[k_1]}$, $\hat{S}_\theta^{[k_2]}$ and $\hat{R}_\theta^{[k_3]}$}
\KwData{$g_\theta$ defined in \eqref{eq:phnn}}
\KwData{Loss function $\mathcal{L}_{g_\theta}$ defined in \eqref{eq:loss_ode}}
\KwResult{Parameters $\theta$ for  $g_\theta$}
\For{k in $1\ldots K$}{
\For{batch in Batches}{
    $B:=\{(u^{j_m}, u^{j_m+1}, t^{j_m})\}_{m=1}^{M_b} \gets \text{DrawRandomBatch}(D, M_b) $\;
    Step using $\mathcal{L}_{g_\theta}(B)$ and $\nabla_\theta \mathcal{L}_{g_\theta}(B)$ 
}
}
\end{algorithm}

\begin{algorithm}
\caption{The training phase of the baseline algorithm}\label{alg:baseline_train}
\KwData{Observations $D=\{(t_1, \vec{x}^1, \vec{u}^1), \ldots, (t_N, \vec{x}^N, \vec{u}^N\}$}

\KwData{Number of epochs $K$}
\KwData{Batch size $M_b$}
\KwData{Initial CNN $g_\theta$}
\KwData{Loss function $\mathcal{L}_{g_\theta}$ defined in \eqref{eq:loss_ode}}
\KwResult{Parameters $\theta$ for  $g_\theta$}
\For{k in $1\ldots K$}{
\For{batch in Batches}{
    $B:=\{(u^{j_m}, u^{j_m+1}, t^{j_m})\}_{m=1}^{M_b} \gets \text{DrawRandomBatch}(D, M_b) $\;
    Step using $\mathcal{L}_{g_\theta}(B)$ and $\nabla_\theta \mathcal{L}_{g_\theta}(B)$ 
}
}
\end{algorithm}
%

\section{Numerical experiments}\label{sec:experiments}

In this section, we test how PHNN models perform on a variety of problems with different properties. To our knowledge, there are no existing methods in the literature for which it is natural to compare the PHNN method across a variety of PDEs. We consider thus first a purely Hamiltonian PDE problem, to be able to compare PHNN to the method of Matsubara et al.\ \cite{Matsubara2020deep}. We also compare our models to the system identification method PDE-FIND  \cite{Rudy2017data} for this problem. For problems with damping and external forces, we have developed our own baseline model that does not have the pseudo-Hamiltonian structure but is otherwise as similar as possible to the PHNNs and is trained in the same way. We test either two or three PHNNs for each problem, in addition to a baseline model. The models we test on all problems are:
\begin{itemize}
    \item PHNN (general): A PHNN model with kernel sizes $k=[3,3,3,1]$ and an $\hat{f}_\theta$ that depends on $u$, $x$ and $t$;
    \item PHNN (informed): A PHNN model where the operators $A$, $S$ and $R$ are known a priori and $\hat{f}_\theta$ depends only on the variable(s) which $f$ depend on;
    \item Baseline: A model consisting of one neural network that takes $u$, $x$ and $t$ as input, where the output is of the same dimension as $u$. The network consists of two parts: first a five-layer deep neural network with a hidden dimension of 20 and the $\tanh$ activation function, then a convolutional layer with kernel size five and activation function $\tanh$, followed by two additional layers with hidden dimension 100 and the $\tanh$ activation function. The first five layers are meant to approximate any non-linear function (say $u\mapsto \frac{1}{2}u^2$ in the case of Burgers' equation), while the convolutional layer is supposed to represent a finite-difference approximation of the spatial derivatives. The baseline model needs a kernel size of five to approximate the third and fourth derivatives present in the first two resp.\ last example we consider.
\end{itemize}

The GitHub repository \url{https://github.com/SINTEF/pseudo-hamiltonian-neural-networks} includes notebooks to run experiments on all the systems we consider in the following. To reproduce the exact results we present in this section and the next, we refer the reader to \url{https://doi.org/10.5281/zenodo.10419436}.

\subsection{The KdV equation}\label{sec:kdv_compare}

If $\nu =0$ in \eqref{eq:kdvburgers}, we get the KdV equation
\begin{equation}\label{eq:kdv}
u_t + \eta u u_x - \gamma^2 u_{xxx} = 0.
\end{equation}
Furthermore, we let $\eta=6$ and $\gamma=1$ and assume periodic solutions $u(0,t) = u(P,t)$ on the domain $[0,P]$, with $P=20$. We generate training data from initial conditions
\begin{equation}\label{eq:kdvinit}
u(x,0) = 2 \sum_{l=1}^2 c_l^2 \operatorname{sech}^2 \bigg( c_l \Big(\big(x + \frac{P}{2} - d_l P\big) \bmod P - \frac{P}{2} \Big) \bigg),
\end{equation}
where $c_1, c_2$ and $d_1, d_2$ are randomly drawn from the uniform distributions $\mathcal{U}(\frac{1}{2}, 2)$ and $\mathcal{U}(0, 1)$ respectively. That is, the initial states are two waves of height $2c_1^2$ and $2c_2^2$ centered at $d_1 P$ and $d_2 P$, with periodicity imposed. The system is integrated from 20 different random initial states from time $t=0$ to time $t=0.2$. This is done with a time step $\Delta t=0.0025$, but then only every fourth step is used as training data. The test data is obtained from 10 random initial states integrated and evaluated at every time step $\Delta t = 0.001$ to $t=2$.

\begin{figure}[ht!]
    \centering
    \includegraphics[width=\textwidth]{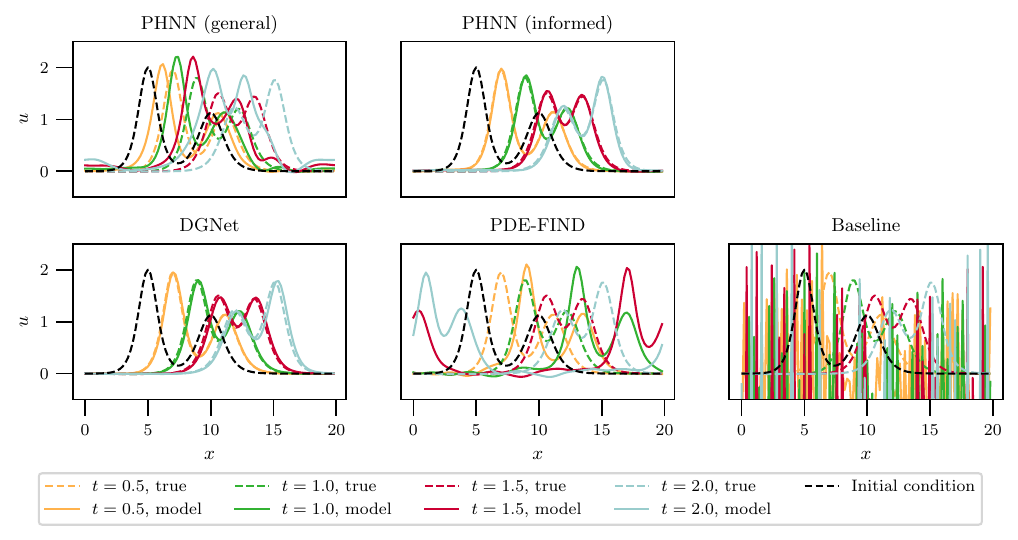}
\caption{Predictions of the KdV equation \eqref{eq:kdv} by two PHNN models and our baseline model, compared to DGNet \cite{Matsubara2020deep} and PDE-FIND \cite{Rudy2017data}. The training data consist of 420 states, with 20 different initial conditions and $21$ points equidistributed in time from $t = 0$ to $t = 0.2$, and the neural network models are all trained for 5000 epochs.}
\label{fig:kdv_compare}
\end{figure}

In this case, where the data actually represents a purely Hamiltonian PDE system, the general PHNN model performs much worse than the informed PHNN model, which in this case becomes a pure HNN model. The DGNet method of Matsubara et al.\ \cite{Matsubara2020deep} can give accurate results, as is evident in Figure \ref{fig:kdv_compare}, but it performs generally worse than our method on varied test data, and also requires much more time to train, as the numbers in Table \ref{tab:kdv_compare} show. The sparse regression method PDE-FIND \cite{Rudy2017data} is not able to find accurate models from the training data considered here.

\begin{table}[ht!]
  \caption{Mean and standard deviation of the MSE from $t=0$ to $t=2$ for 10 models of each type tested on 10 different initial conditions, compared to the numerical solution of the exact KdV equation \eqref{eq:kdv}. The runtime provided is the average training time in seconds for each method, on two $2.4$ GHz CPU cores of Intel Xeon Gold 6126.}
  \centering
        \begin{tabular}{lrrrrrr}
        \toprule
        & \multicolumn{3}{c}{$5000$ epochs} & \multicolumn{3}{c}{$20000$ epochs} \\
        & mean MSE & std MSE & Runtime & mean MSE & std MSE & Runtime \\
        \midrule
            PHNN (general) & 1.75e+02 & 3.62e+02 & $2691$ & 6.52e+01 & 9.27e+01 & $10603$ \\
            PHNN (informed) & 1.34e+00 & 5.96e+00 & $1172$ & 7.23e-01 & 3.70e+00 & $4587$ \\
            DGNet \cite{Matsubara2020deep} & 4.58e+01 & 2.45e+02 & $9555$ & 3.45e+01 & 1.99e+02 & $37227$ \\
            PDE-FIND \cite{Rudy2017data} & Inf & Inf & $2$ & Inf & Inf & $2$ \\
            Baseline & 1.34e+03 & 3.67e+03 & $1022$ & 1.03e+03 & 2.20e+03 & $4245$ \\
        \bottomrule
        \end{tabular}
  \label{tab:kdv_compare}
\end{table}

\subsection{The KdV--Burgers equation}\label{sec:kdv}
Consider now the KdV--Burgers equation from examples \ref{ex:kdvburgers} and \ref{ex:kdvburgersdisc}, but with external forces:
\begin{equation}\label{eq:forcedkdvburgers}
u_t + \eta (\frac{1}{2}u^2)_x - \nu u_{xx} - \gamma^2 u_{xxx} = f(x,t).
\end{equation}
The spatial domain is still $[0,P]$ with $u(0,t) = u(P,t)$ and $P=20$. We let $\eta=6$, $\nu=0.3$, $\gamma=1$ and $f(x,t) = \sin{\left(\frac{2\pi x}{P}\right)}$, and generate again training data from random initial conditions \eqref{eq:kdvinit}.

We compare the general and informed PHNN models to the baseline model on \eqref{eq:forcedkdvburgers} with
\begin{equation}\label{eq:kdvf}
    f(x,t) = \frac{3}{5}\sin{\big( \frac{4\pi}{P}x -t\big)}.
\end{equation}
Ten models of each type are trained using training sets consisting of $410$ states, obtained from integrating $10$  random initial states of the form \eqref{eq:kdvinit} and evaluating the solution at every time step $\Delta t=0.05$ until $t=2$. We train on a larger time domain here than we did for the KdV equation to learn the explicit dependence of $f$ on $t$.

\begin{figure}[ht!]
    \centering
    \includegraphics[width=\textwidth]{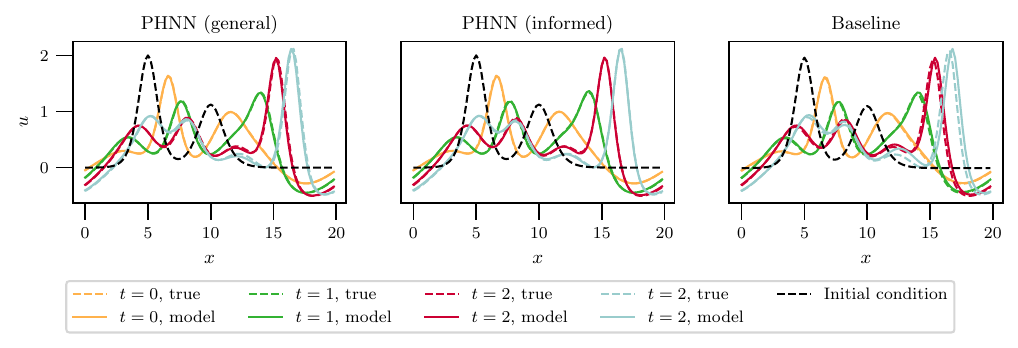}
\caption{Predictions of the forced KdV--Burgers system \eqref{eq:forcedkdvburgers} with force \eqref{eq:kdvf}, obtained from the best of 10 models of each model type, after being trained for 5000 epochs, as evaluated by the mean MSE at $t=2$ on predictions from 10 random initial states.}
\label{fig:kdv_time_5000}
\end{figure}

\begin{figure}[ht!]
    \centering
    \includegraphics[width=\textwidth]{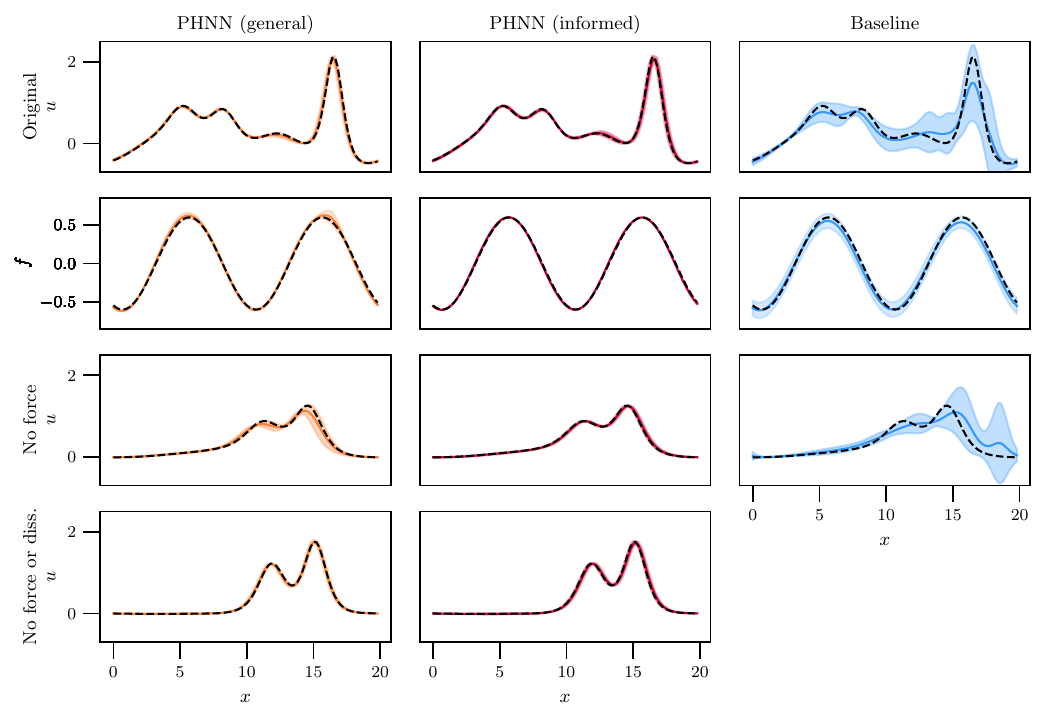}
\caption{Solutions of the various models, after being trained for 5000 epochs, of the forced KdV--Burgers system \eqref{eq:forcedkdvburgers} with $f$ given by \eqref{eq:kdvf} at time $t=2$. The line and the shaded area are the mean resp.\ standard deviation of 10 models of each type. The dashed black line is the ground truth. \textit{Upper row:} The original system \eqref{eq:forcedkdvburgers} that the models are trained on. \textit{Second row:} The learned force approximating $f$ in \eqref{eq:forcedkdvburgers}. \textit{Third row:} Predictions with the force $f$ removed from the models. \textit{Lower row:} Predictions with the external force and the dissipation term removed from the models.}
\label{fig:kdv_std_5000}
\end{figure}

Even when the baseline model is able to approximate the dynamics well, the training of the model generally converges more slowly than the PHNN models. After training 10 models of each type for 5000 epochs, the PHNN models are consistently outperforming the baseline model; see figures \ref{fig:kdv_time_5000} and \ref{fig:kdv_std_5000}.

\begin{figure}[ht!]
    \centering
    \includegraphics[width=\textwidth]{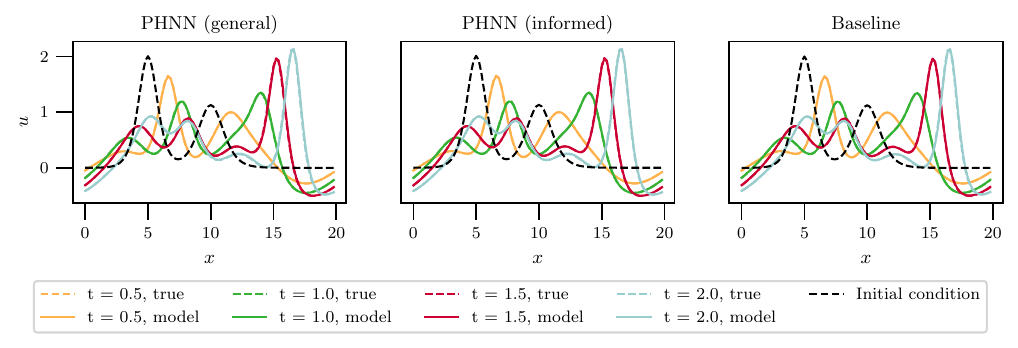}
\caption{Predictions of the forced KdV--Burgers system \eqref{eq:forcedkdvburgers} with force \eqref{eq:kdvf}, obtained from the best of 10 models of each model type, as evaluated by the mean MSE at $t=2$ on predictions from 10 random initial states.}
\label{fig:kdv_time}
\end{figure}

\begin{figure}[ht!]
    \centering
    \includegraphics[width=\textwidth]{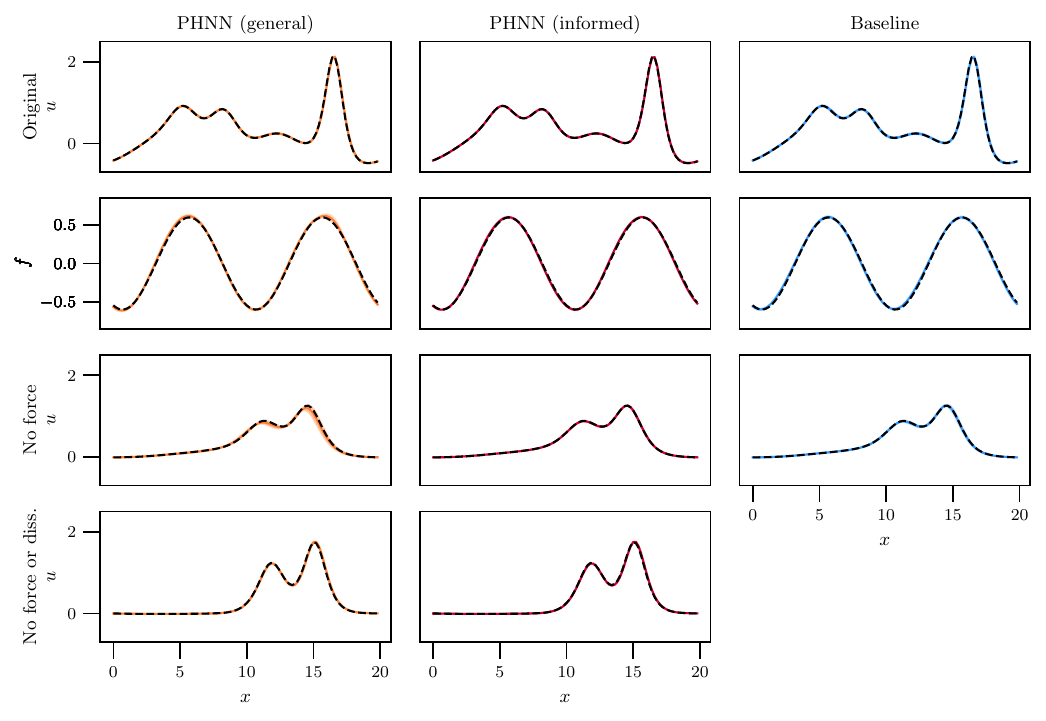}
\caption{Solutions of the various learned models of the forced KdV--Burgers system \eqref{eq:forcedkdvburgers} with $f$ given by \eqref{eq:kdvf} at time $t=2$. The line and the shaded area, barely visible in these plots, are the mean resp.\ standard deviation of 10 models of each type. The dashed black line is the ground truth. \textit{Upper row:} The original system \eqref{eq:forcedkdvburgers} that the models are trained on. \textit{Second row:} The learned force approximating $f$ in \eqref{eq:forcedkdvburgers}. \textit{Third row:} Predictions with the force $f$ removed from the models. \textit{Lower row:} Predictions with the external force and the dissipation term removed from the models.}
\label{fig:kdv_std}
\end{figure}

To test the accuracy of the models after the training has converged, we then train the PHNN models for 20 000 epochs and the baseline models for 50 000 epochs. At every epoch, the models are validated by integrating them to time $t=2$ starting at three random initial states and calculating the average mean squared error (MSE) from these. The model with the lowest validation score after the last epoch is saved as the final model. When being tested on an initial state well within the domain the training data is sampled from, all models perform well; see figures \ref{fig:kdv_time} and \ref{fig:kdv_std}.

When they are tested on a wide range of initial states, some of the models struggle to give stable and accurate solutions. We observe that the PHNN models are quite sensitive to variations in the initialization of the learnable parameters of the model, an issue we discuss further in Section \ref{sec:stability}. Hence we get a large average MSE from these models when applying them on 10 different initial states, as is evident from Table \ref{tab:kdv}. However, the best PHNN models perform well on all the test cases; of the 30 models trained, 10 of each type, the seven models with the lowest average MSE on 10 test sets are all PHNN models. Thus it would be advisable to run several PHNN models with different initalizations of the neural networks and disregard those models who behave vastly different from the others. We demonstrate this by picking out the three most similar models of each type, as measured by their predictions on 10 different random initial conditions, and evaluate the mean error on those; see Table \ref{tab:kdv}.

\begin{table}[ht!]
  \caption{Mean and standard deviation of the MSE at $t=2$, for 10 models of each type and for the three most similar of each type, trained on the KdV--Burgers equation with the external force \eqref{eq:kdvf}.}
  \centering
    \begin{tabular}{lrrrr}
    \toprule
            & \multicolumn{2}{c}{10 models} & \multicolumn{2}{c}{Three models} \\
              & \multicolumn{1}{c}{mean} & \multicolumn{1}{c}{std} & \multicolumn{1}{c}{mean} & \multicolumn{1}{c}{std} \\
    \midrule
        PHNN (general) & 1.00e+01 & 1.42e+01 & 4.32e-01 & 6.09e-03 \\
        PHNN (informed) & 3.42e+01 & 3.44e+01 & 4.25e-01 & 6.73e-03 \\
        Baseline & 2.23e+00 & 1.61e+00 & 5.82e-01 & 1.53e-01 \\
    \bottomrule
    \end{tabular}
  \label{tab:kdv}
\end{table}


Figures \ref{fig:kdv_std_5000} and \ref{fig:kdv_std} demonstrate one of the main qualitative features of the PHNN models: we can remove the force and dissipation from the model and still get an accurate solution of the system without these. In these figures, we have also extracted the external force part from the baseline model by
\begin{equation}
    \hat{f}_\theta(x,t) := \hat{g}_\theta(u,x,t) - \hat{g}_\theta(u,0,0).
\end{equation}
This works here, since the external force is independent of the solution states and the integrals are zero when $u=0$, but it would not be an option in general. 
Moreover, we note that there is no way to separate the conservation and dissipation terms of the baseline model. This can however be done with the PHNN models, so that we also have a model for the energy-preserving KdV equation.

\begin{figure}[ht!]
    \centering
    \includegraphics[width=\textwidth]{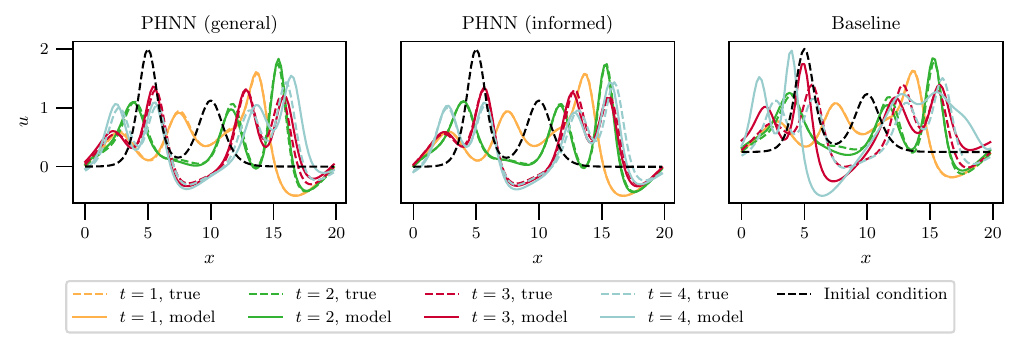}
\caption{Predictions of the forced KdV--Burgers system \eqref{eq:forcedkdvburgers} with $f$ given by \eqref{eq:kdvfx} obtained from the best of 10 models of each model type, as evaluated by the mean MSE at $t=0.5$ on predictions from 10 random initial states.}
\label{fig:kdv_time_fx}
\end{figure}

\begin{figure}[ht!]
    \centering
    \includegraphics[width=\textwidth]{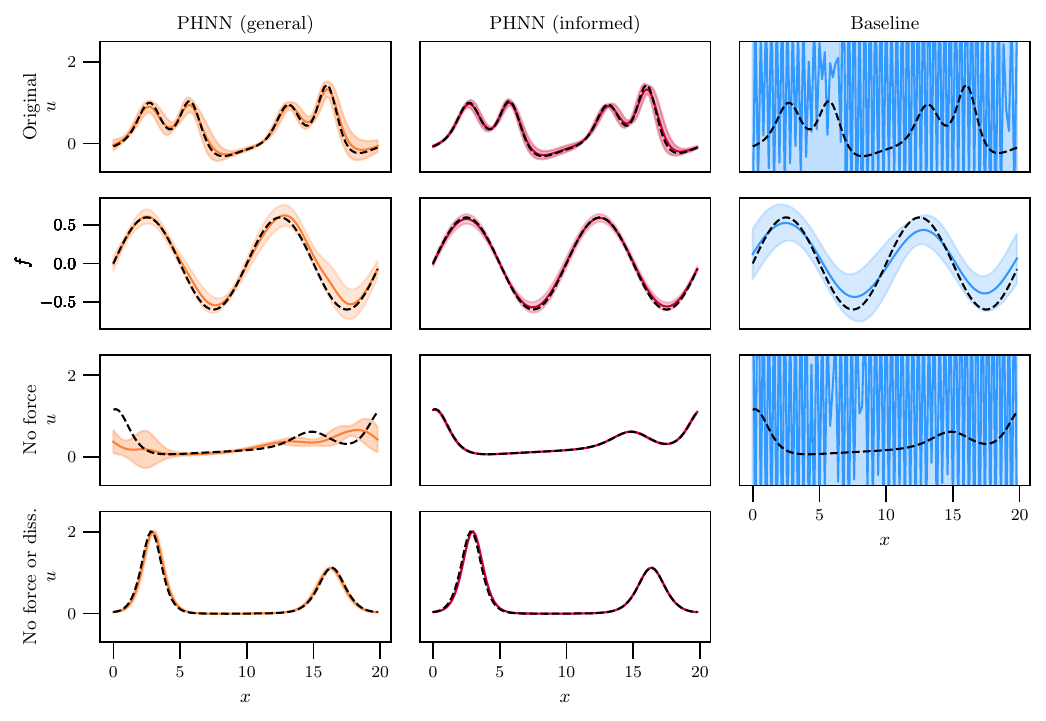}
\caption{Solutions of the various learned models of the forced KdV--Burgers system with $f$ given by \eqref{eq:kdvfx}. The line and the shaded area is the mean resp.\ standard deviation of predictions at $t=4$ of 10 models of each type. The dashed black line is the ground truth. \textit{Upper row:} The original system. \textit{Second row:} The learned force approximating $f$ in \eqref{eq:kdvfx}. \textit{Third row:} Predictions with the external force $f$ removed from the models. \textit{Lower row:} Predictions with the force and the dissipation term removed.}
\label{fig:kdv_std_fx}
\end{figure}

When the external force is explicitly dependent on time, we are generally not able to learn a model that is accurate beyond the temporal domain of the training data. For autonomous systems, we can make due with less training data. Consider thus instead of \eqref{eq:kdvf} the external force
\begin{equation}\label{eq:kdvfx}
    f(x) = \frac{3}{5}\sin{\big( \frac{4\pi}{P}x\big)}.
\end{equation}
We train models for \eqref{eq:forcedkdvburgers} with this $f$, where now we have training sets consisting of $60$ states, from solutions obtained at every time step $\Delta t=0.1$ from $t=0$ to $t=0.5$. As can be seen in figures \ref{fig:kdv_time_fx} and \ref{fig:kdv_std_fx}, the PHNN models perform better than the baseline model, and especially so with increasing time. The worst-performing baseline models become unstable before the final test time $t=4$. We also see that the most general PHNN model struggles to correctly separate the external force from the viscosity term with this amount of training data. However, this is not an issue when the model is informed that the force is purely dependent on the spatial variable.

\subsection{The forced BBM equation}
The Benjamin--Bona--Mahony (BBM) equation was introduced as an improvement on the KdV equation for modelling waves on a shallow surface \cite{Peregrine1966calculations, Benjamin1972model}. We consider this equation with a time- and state-dependent source term:
\begin{equation}\label{eq:bbm}
u_t - u_{xxt} + u_x + u u_x = f(u, t),
\end{equation}
which can be written on the form \eqref{eq:pdeclassspef} with $A = 1- \frac{\partial^2}{\partial x^2}$, $S = \frac{\partial}{\partial x}$ and $R=0$. This requires
\begin{equation*}
\mathcal{H} = \frac{1}{2} \int_\Omega \Big( u^2 + \frac{1}{3}u^3 \Big) \, \de x.
\end{equation*}

As for the KdV--Burgers equation, we train the model on a forced system starting with a two-soliton initial condition. In this case, the initial states are given by
\begin{equation}\label{eq:bbminit}
u(x,0) = 3 \sum_{l=1}^2 (c_l-1) \operatorname{sech}^2 \bigg( \frac{1}{2}\sqrt{1-\frac{1}{c_l}} \Big(\big(x + \frac{P}{2} - d_l P\big) \bmod P - \frac{P}{2} \Big) \bigg),
\end{equation}
i.e.\ two waves of amplitude $3(c_1-1)$ and $3(c_2-1)$ centered at $d_1 P$ and $d_2 P$, where $c_1, c_2$ and $d_1, d_2$ are randomly drawn from the uniform distributions $\mathcal{U}(1, 4)$ and $\mathcal{U}(0, 1)$ respectively, and with periodicity imposed on $\Omega = [0,P]$. We set $P=50$ in the numerical experiments. Furthermore, we let
\begin{equation}\label{eq:ch_fut}
    f(u,t) = \frac{1}{10} \sin(t) u.
\end{equation}

In addition to the three models described in the introduction of this section, we also test a model that is identical to the most general PHNN model except that it does not include a dissipation term. We do this because it is not clearly defined whether or how \eqref{eq:ch_fut} should be separated into a term that is constantly dissipative and one that is not. The most general model does learn that the system has a non-zero dissipative term; however, this term added to the learned force term is close to the ground truth force term. This is due to a leakage of a term $\alpha u$ for some random constant $\alpha$ between the terms, similar to the constant leakage described in Section \ref{sec:leakage}, so that we learn an approximated integral $\hat{\mathcal{V}}_\theta = \alpha \frac{\Delta x}{2} \sum_{i=0}^M u_i^2$ with $\hat{R}_\theta^{[3]} = I$ and a corresponding external force $\hat{f}_\theta = (\frac{1}{20}\sin{(t)}+\alpha) u$. This leakage could be combatted by regularization, i.e.\ by penalizing the mean absolute value of the dissipation term. We do not do that in the numerical experiments presented here, but instead opt to also learn a model without the dissipation term and compare to this.

\begin{figure}[ht!]
    \centering
    \includegraphics[width=\textwidth]{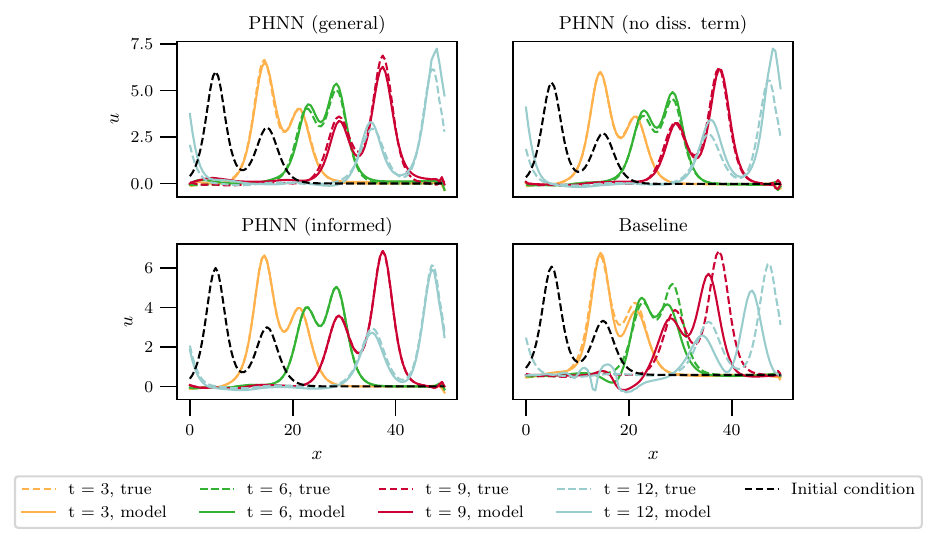}
\caption{Predictions of the forced BBM system \eqref{eq:bbm} obtained from the best of 10 models of each model type, as evaluated by the mean MSE at $t=10$ on predictions from 10 random initial states.}
\label{fig:bbm_time_alt}
\end{figure}

\begin{figure}[ht!]
    \centering
    \includegraphics[width=\textwidth]{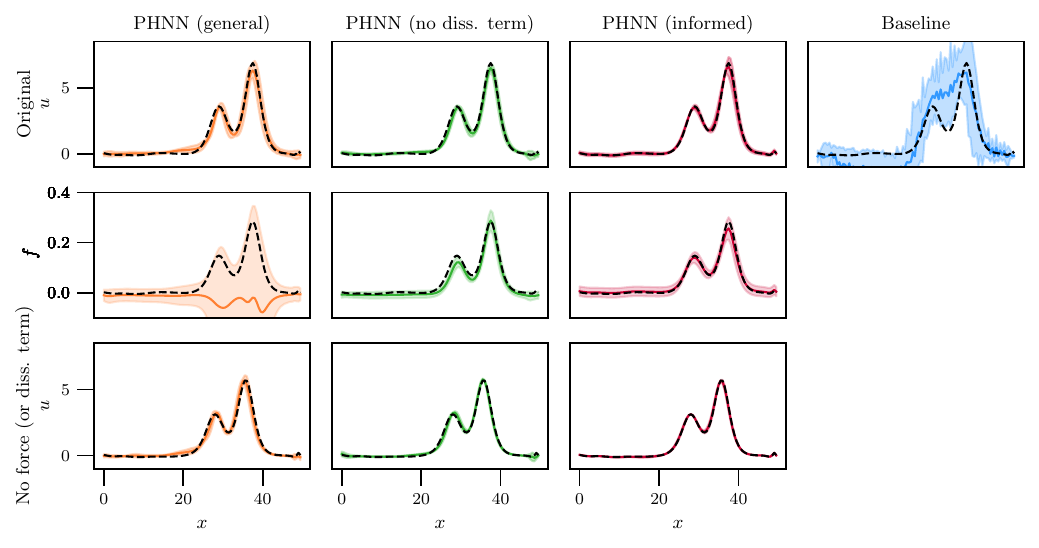}
\caption{Solutions of the learned BBM system obtained from the different models. The line and the shaded area is the mean resp.\ standard deviation of predictions at $t=9$ of 10 models of each type. The dashed black line is the ground truth. \textit{Upper row:} The original system \eqref{eq:bbm} that the models are trained on. \textit{Middle row:} The learned force approximating $f$ in \eqref{eq:bbm}. \textit{Lower row:} Predictions with the force $f$ removed from the models.}
\label{fig:bbm}
\end{figure}

For every type of model, we train 10 distinct models with random initializations for a total of 20 000 epochs for the PHNN models and 50 000 epochs for the baseline model. We use training data comprising $260$ states, obtained from integrating $10$ randomly drawn initial states with time step $\Delta t=0.4$ from time $t=0$ to time $t=10$. A validation score at each epoch is generated by integrating the models to time $t=1$ starting at three initial states and calculating the mean MSE, and then the model with the lowest validation score is kept. After training is done, we integrate the models starting from 10 new arbitrary initial conditions to determine the average MSE at $t=10$. By the error of the models as reported in Table \ref{tab:bbm}, we see that all PHNN models perform better than the baseline model. Interestingly, the average MSE is lowest for the PHNN model where $A$ and $S$ has to be learned but $R$ is known to be zero. However, the best PHNN model of all 30 models trained is one of those informed of these operators, as seen in Figure \ref{fig:bbm_time_alt}. Note that the baseline model cannot be expected to learn a perfect model for this example, since the discrete approximation of $A^{-1} S = (1- \frac{\partial^2}{\partial x^2})^{-1}\frac{\partial}{\partial x}$ used when generating the training data is a discrete convolution operator with kernel size bigger than five. The PHNN models are more stable and behave especially well with increasing time compared to the baseline model. Beyond $t=10$, the accuracy of all models quickly deteriorates. We attribute this to the poor extrapolation abilities of neural networks; the models are not able to learn how the time-dependent $f$ behaves beyond the temporal domain $t\in[0,10]$ of the training data. Figure \ref{fig:bbm} shows the external forces learned by the PHNNs, and how well the models predict the system with these forces removed. For the general PHNN model, we have in this case also removed the dissipation term.

\begin{table}[ht!]
  \caption{Mean and standard deviation of the MSE at $t=10$, for 10 models of each type and for the three most similar of each type, trained on the BBM equation with an external force.}
  \centering
    \begin{tabular}{lrrrr}
    \toprule
            & \multicolumn{2}{c}{10 models} & \multicolumn{2}{c}{Three models} \\
               & \multicolumn{1}{c}{mean} & \multicolumn{1}{c}{std} & \multicolumn{1}{c}{mean} & \multicolumn{1}{c}{std} \\
    \midrule
        PHNN (general) & 6.13e-01 & 8.04e-01 & 1.95e-01 & 1.02e-01\\
        PHNN (no diss. term) & 1.19e-01 & 7.75e-02 & 4.79e-02 & 2.55e-02  \\
        PHNN (informed) & 1.46e-01 & 4.10e-01 & 3.45e-03 & 3.33e-03 \\
        Baseline & 4.81e+00 & 4.79e+00 & 8.61e-01 & 3.29e-01 \\
    \bottomrule
    \end{tabular}
  \label{tab:bbm}
\end{table}

\subsection{The Perona--Malik equation}\label{sec:pm}
In addition to modelling physical systems, PDEs can be used for image restoration and denoising. For instance, if the heat equation is applied to a greyscale digital image, where the state $u$ gives the intensity of each pixel, it will smooth out the image with increasing time. The Perona--Malik equation for so-called anisotropic diffusion is designed to smooth out noise but not the edges of an image \cite{Perona1990scale}. Several variations of the equation exist. We consider the one-dimensional case, with a space-dependent force term, given by
\begin{equation}\label{eq:pm}
    u_t + \left( \frac{u_x}{1+u_x^2}\right)_x = f(x).
\end{equation}
This is a PDE of the type \eqref{eq:pdeclassspef} with $A = I$, $S=0$ and $R=I$, and
\begin{equation*}
    \mathcal{V}[u] = \frac{1}{2} \int_\Omega\ln(1 + u_x^2) \, \de x.
\end{equation*}
Note that the equation can be written on the form $u_t = \frac{\partial}{\partial x} \phi[u] + f(x)$ for $\phi[u] = -\frac{u_x}{1+u_x^2}$, but this $\phi[u]$ is not the variational derivative of any integral. We consider \eqref{eq:pm} on the domain $[0,P]$ with $P=6$, and set
\begin{equation}\label{eq:pm_force}
f(x) = 10 \sin{\big(\frac{4\pi}{P}x\big)}
\end{equation}
for the following experiments. The initial conditions are given by
\begin{equation}\label{eq:pm_init}
u(x,0) = a - \sum_{l=1}^2 \Big( h_l \Big( \tanh{\big(b (x-d_l)\big)} - \tanh{\big(b(x-P+d_l)\big)} \Big)\Big) + c \sin^2{(r \pi x)} \sin{(s \pi x)}
\end{equation}
where $a \in \mathcal{U}(-5,5)$, $b\in \mathcal{U}(20,40)$, $c \in \mathcal{U}(0.05,0.15)$, $d_l\in \mathcal{U}(0.3,3)$, $h_l\in \mathcal{U}(0.5,1.5)$, $r \in \mathcal{U}(0.5,3)$ and $s \in \mathcal{U}(10,20)$.

\begin{figure}[ht!]
    \centering
    \includegraphics[width=\textwidth]{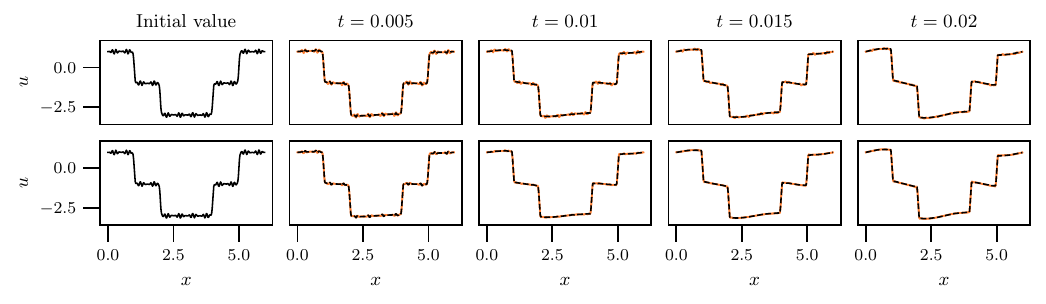}
\caption{The result at different times from integrating the mean of five general PHNN models trained on the Perona--Malik system \eqref{eq:pm} using two different integration schemes in the training. The solution of the learned models is in yellow, while the dashed black line is the solution of the exact PDE. \textit{Upper row:} The second-order midpoint method. \textit{Lower row:} The fourth-order symmetric method SRK4.}
\label{fig:pm_integrators}
\end{figure}

We train 10 models of each type for 10 000 epochs, on 20 pairs of data at time $t=0$ and $t=0.02$. This corresponds to the original noisy image and an image where the noise is almost completely removed, as judged by visual inspection. Because the step between these states is quite large, a high-order integrator is required to get an accurate approximation of the time-derivative. Indeed, models trained with the second-order implicit midpoint method fail to remove the noise as fast or accurately as the ground truth \eqref{eq:pm}. Thus we use instead the fourth-order symmetric Runge--Kutta method (SRK4) introduced in \cite{Eidnes2023pseudo}. This requires roughly four times the computational cost per epoch as using the midpoint method, but gives a considerably improved performance, as demonstrated in Figure \ref{fig:pm_integrators}.

\begin{figure}[ht!]
    \centering
    \includegraphics[width=\textwidth]{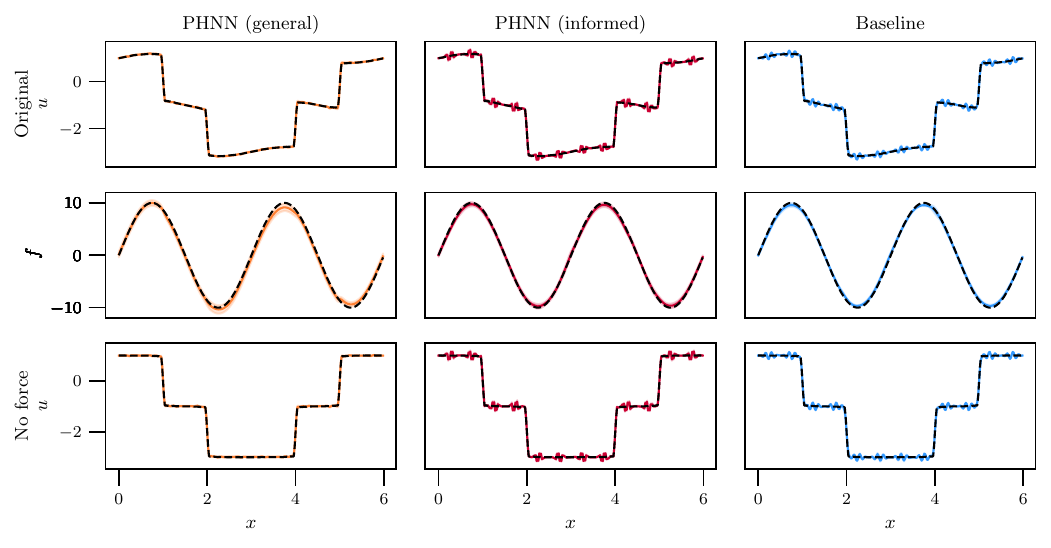}
\caption{Perona--Malik at time $t=0.02$, models and exact. The line plot is the average of 10 models of each type, while the shaded region indicates the standard deviation. \textit{Upper row:} The original system \eqref{eq:ch} that the models are trained on. \textit{Middle row:} The learned force approximating \eqref{eq:pm_force}. \textit{Lower row:} Predictions with the force $f$ removed from the models.}
\label{fig:pm}
\end{figure}

Table \ref{tab:pm} and Figure \ref{fig:pm} report the result of applying the learned models on an original noisy state \eqref{eq:pm_init} with $a=1$, $b=30$, $c=0.15$, $d_1=1$, $d_2=2$, $h_1=h_2=1$, $r=2$ and $s=15$. 
Interestingly, the general PHNN model performs better than the informed one. Moreover, the PHNN models perform better when the kernel size of the first convolutional layer of $\hat{\mathcal{V}}_\theta$ is three instead of two. This indicates that the model does not learn the Perona--Malik equation but rather a different PDE that denoises the image. This may be as expected when we only train on initial states and end states. An odd-numbered filter size is the norm when convolutional neural networks are used for imaging tasks, since this helps to maintain spatial symmetry, and the improved performance with a kernel of size three in $\hat{\mathcal{V}}_\theta$  can perhaps be related to this.

\begin{table}[ht!]
  \caption{Mean and standard deviation of the MSE at end time $t=0.02$, for 10 models of each type and for the three most similar of each type, trained on the Perona--Malik equation with an external force, and evaluated on predictions from 10 random initial states.}
  \centering
    \begin{tabular}{lrrrr}
        \toprule
                & \multicolumn{2}{c}{10 models} & \multicolumn{2}{c}{Three models} \\
                  & \multicolumn{1}{c}{mean} & \multicolumn{1}{c}{std} & \multicolumn{1}{c}{mean} & \multicolumn{1}{c}{std} \\
        \midrule
            PHNN (general) & 5.36e-04 & 1.58e-04 & 4.01e-04 & 3.98e-05 \\
            PHNN (informed) & 4.14e-03 & 1.03e-03 & 3.85e-03 & 5.03e-04 \\
            Baseline & 3.09e-03 & 3.08e-04 & 3.04e-03 & 4.01e-05 \\
        \bottomrule
    \end{tabular}
  \label{tab:pm}
\end{table}

\subsection{The Cahn--Hilliard equation}\label{sec:ch}
The Cahn--Hilliard equation was originally developed for describing phase separation \cite{Cahn1958free}, but has applications also in image analysis, and specifically image inpainting \cite{Burger2009cahn, Schoenlieb2015partial}. Machine learning of pattern-forming PDEs, which include the Cahn--Hilliard and Allen--Cahn equations, has been studied in \cite{Zhao2020learning}. Results on applying PHNN to the Allen--Cahn equation is included in our GitHub repository. However, here we only consider the Cahn--Hilliard equation, with an external force, given by
\begin{equation}\label{eq:ch}
u_t  - (\nu u + \alpha u^3 + \mu u_{xx})_{xx} = f(u,x).
\end{equation}
This is a dissipative PDE if the external force is zero, and it can be written on the form \eqref{eq:pdeclassspef} with $A = I$, $S=0$ and $R=-\frac{\partial^2}{\partial x^2}$, and
\begin{equation*}
\mathcal{V}[u] = \frac{1}{2} \int_\Omega \left(\nu u^2 + \frac{1}{2}\alpha u^4  - \mu u_x^2\right)\, dx.
\end{equation*}
In the experiments, we set $\nu=-1$, $\alpha= 1$ and $\mu=-\frac{1}{1000}$, and
\begin{equation*}
    f(u,x) =
    \begin{cases}
        30 u & \text{if } 0.3 < x < 0.7,\\
        0              & \text{otherwise}.
    \end{cases}
\end{equation*}
The initial conditions of the training data are 
\begin{equation}\label{eq:ch_init}
    u(x,0) = \sum_{l=1}^2 \Big(a_l \sin{\big(c_l \frac{2 \pi }{P} x\big)} + b_l \cos{\big(d_l \frac{2 \pi}{P} x\big)}\Big)
\end{equation}
on the domain $[0,P]$ with $P=1$, where $a_l$, $b_l$, $c_l$ and $d_l$ are random parameters from the uniform distributions $\mathcal{U}(0,\frac{1}{5})$, $\mathcal{U}(0,\frac{1}{20})$, $\mathcal{U}(1,6)$ and $\mathcal{U}(1,6)$, respectively.

\begin{figure}[ht!]
    \centering
    \includegraphics[width=\textwidth]{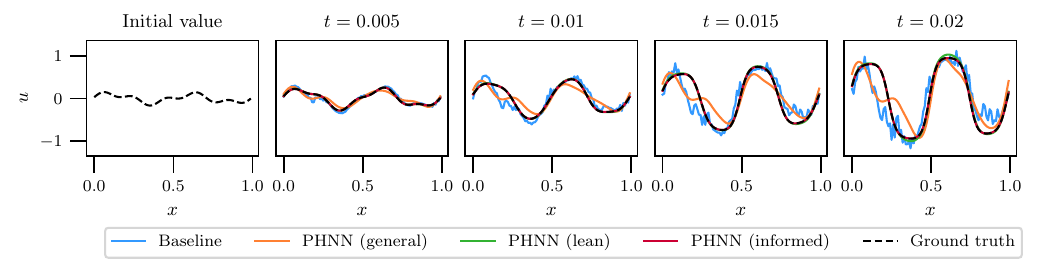}
\caption{Predictions of the forced Cahn--Hilliard system obtained from the best of 10 models of each model type, as evaluated by the mean MSE at $t=0.02$ on predictions from 10 random initial states.}
\label{fig:ch_time}
\end{figure}

\begin{figure}[ht!]
    \centering
    \includegraphics[width=\textwidth]{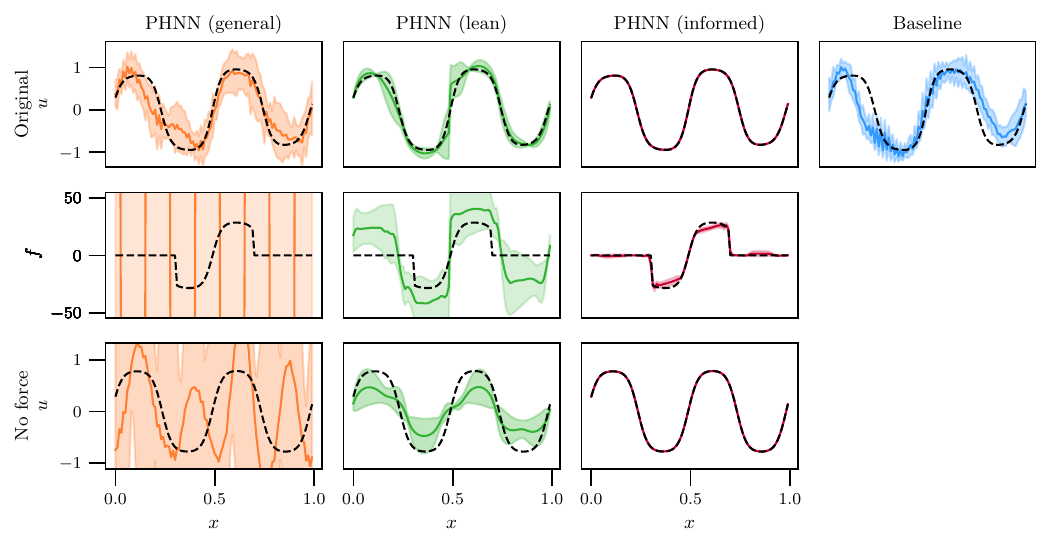}
\caption{Mean and standard deviation of predictions at $t=0.02$ obtained from integrating 10 models of each type, for the Cahn--Hilliard problem \eqref{eq:ch}. The dashed black line is the ground truth. \textit{Upper row:} The original system \eqref{eq:ch} that the models are trained on. \textit{Middle row:} The learned force approximating $f$ in \eqref{eq:ch}. \textit{Lower row:} Predictions with the force $f$ removed from the models.}
\label{fig:ch}
\end{figure}

In addition to the models described in the introduction of this section, we also train a "lean" model, with $k=[1,0,3,1]$ but no prior knowledge of how $R$ looks. For each model type, 10 randomly initialized models are trained for 20 000 (for the PHNN models) or 50 000 (for the baseline models) epochs on different randomly drawn data sets consisting of a total of 300 states, at times $t=0$, $t=0.004$ and $t=0.008$. At each epoch, the model is evaluated by comparing to the ground truth solution of three states at $t=0.008$, and the model with the lowest MSE on this validation set is kept. The resulting 10 models are then evaluated on 10 random initial conditions and the mean MSE in the last time step is calculated from this. The mean and standard deviation from all 10 models of each type, and the three most similar of each type, are given in Table \ref{tab:ch}. The prediction of the model of each type with the lowest mean MSE is shown in Figure \ref{fig:ch_time}. In Figure \ref{fig:ch} we give the results of the average of all models, and the standard deviation. Here we also show the learned external force and the prediction when this is removed from the model. The initial state of the plots in figures \ref{fig:ch_time} and \ref{fig:ch} is \eqref{eq:ch_init} with $a_1=0.1$, $a_2=0.06$, $b_1=0.01$, $b_2=0.02$, $c_1=2$, $c_2=5$, $d_1=1$, $b_2=2$.

We see from Figure \ref{fig:ch_time} that the most general PHNN model may model the system moderately well, but it is highly sensitive to variations in the training data and the initialization of the neural networks in the model; from Figure \eqref{fig:ch} and Table \ref{tab:ch} we see that this model may produce unstable predictions. In any case, the PHNN models struggle to learn the external force of this problem accurately without knowing $R$, which we see by comparing the predictions of \textit{PHNN (lean)} and \textit{PHNN (informed)} in Figure \ref{fig:ch}, where the difference between the models is that the former has to learn an approximation $\hat{R}_\theta^{[2]}$ of $R$ and is not informed that $f$ is not explicitly time-dependent.

\begin{table}[ht!]
  \caption{Mean and standard deviation of the MSE at $t=0.02$, for 10 models of each type and for the three most similar of each type, trained on the forced Cahn--Hilliard system \eqref{eq:ch} and evaluated on predictions from 10 random initial states.}
  \centering
    \begin{tabular}{lrrrr}
        \toprule
                & \multicolumn{2}{c}{10 models} & \multicolumn{2}{c}{Three models} \\
                   & \multicolumn{1}{c}{mean} & \multicolumn{1}{c}{std} & \multicolumn{1}{c}{mean} & \multicolumn{1}{c}{std} \\
        \midrule
            PHNN (general) & 1.14e+00 & 8.70e-01 & 4.61e-01 & 2.57e-01 \\
            PHNN (lean) & 2.69e-01 & 1.36e-01 & 2.12e-01 & 3.42e-02 \\
            PHNN (informed) & 2.49e-02 & 2.88e-04 & 2.48e-02 & 6.46e-05 \\
            Baseline & 2.77e-01 & 7.75e-02 & 1.99e-01 & 2.24e-02 \\
        \bottomrule
\end{tabular}
  \label{tab:ch}
\end{table}

\section{Analysis of the models and further work}
Here we provide some preliminary analysis of the PHNN models, which lays the groundwork for further analysis and development to be performed in the future.

\subsection{Stability with respect to initial neural network}\label{sec:stability}
\begin{figure}[ht!]
    \centering
    \begin{subfigure}{0.48\textwidth}
    \includegraphics[width=\textwidth]{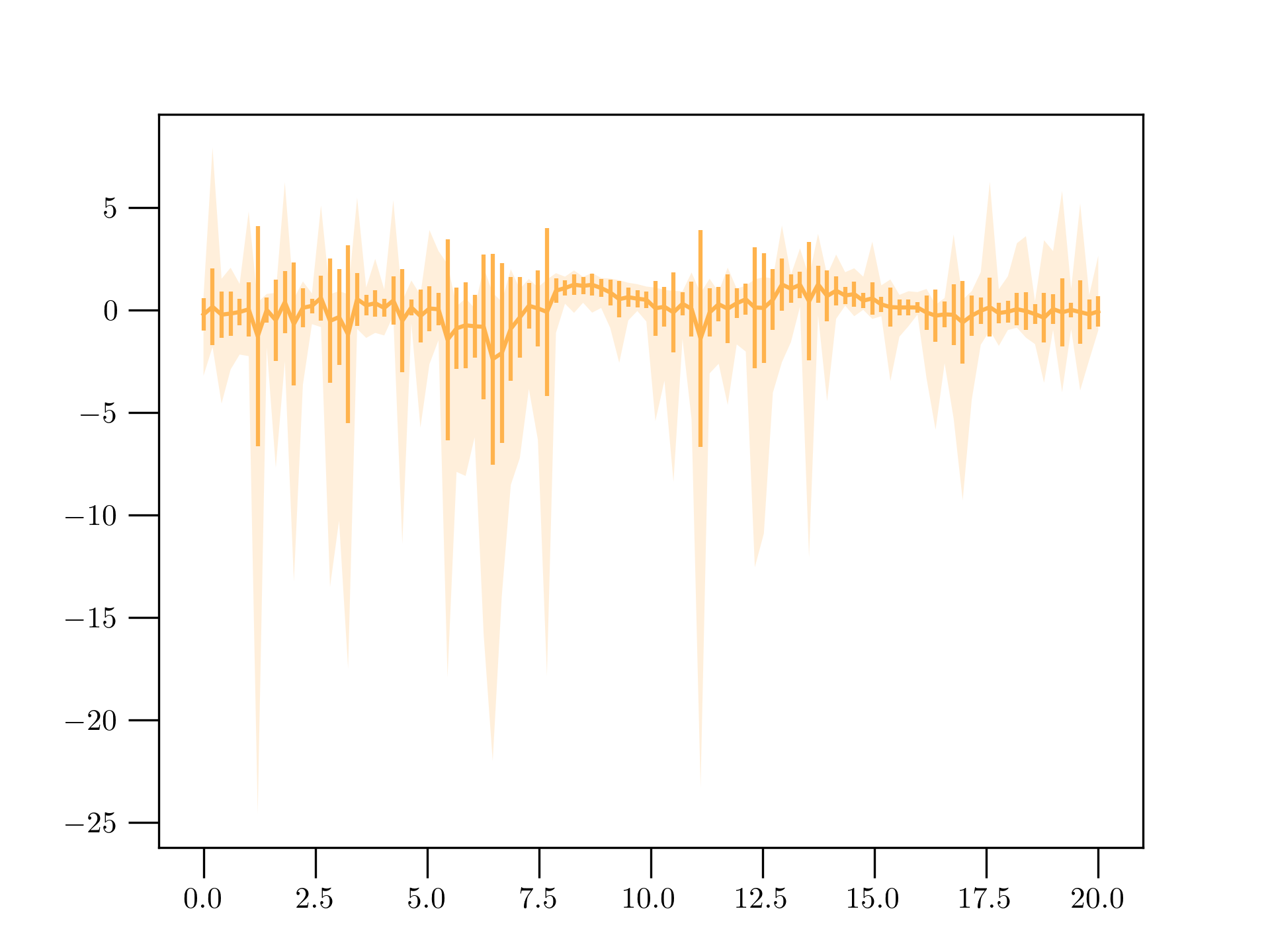}
    \caption{Baseline, 1000 epochs, $T=1$.}
    \end{subfigure}
    \begin{subfigure}{0.48\textwidth}
    \includegraphics[width=\textwidth]{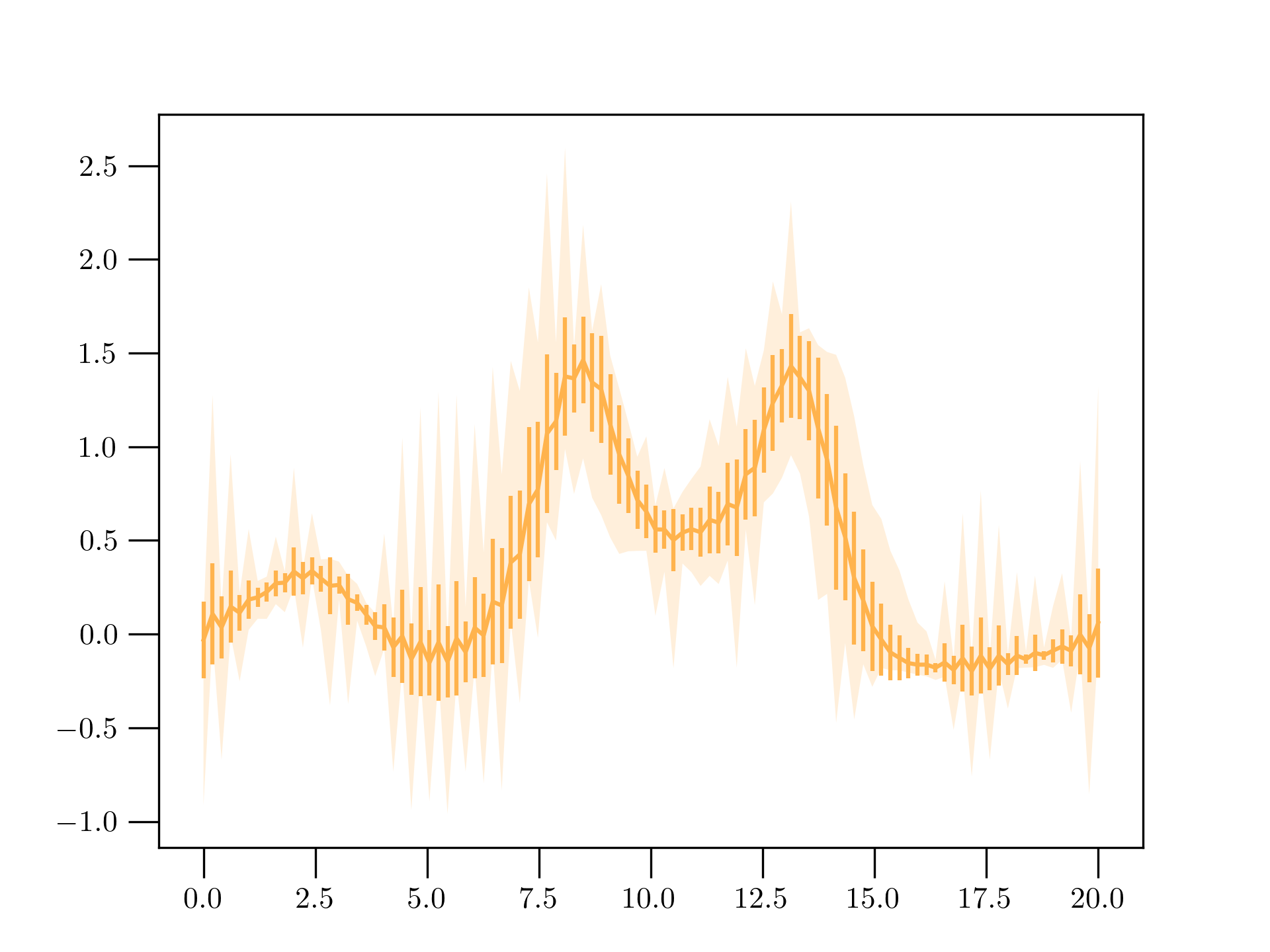}
    \caption{PHNN, 1000 epochs, $T=1$.}
    \end{subfigure}

    \begin{subfigure}{0.48\textwidth}
    \includegraphics[width=\textwidth]{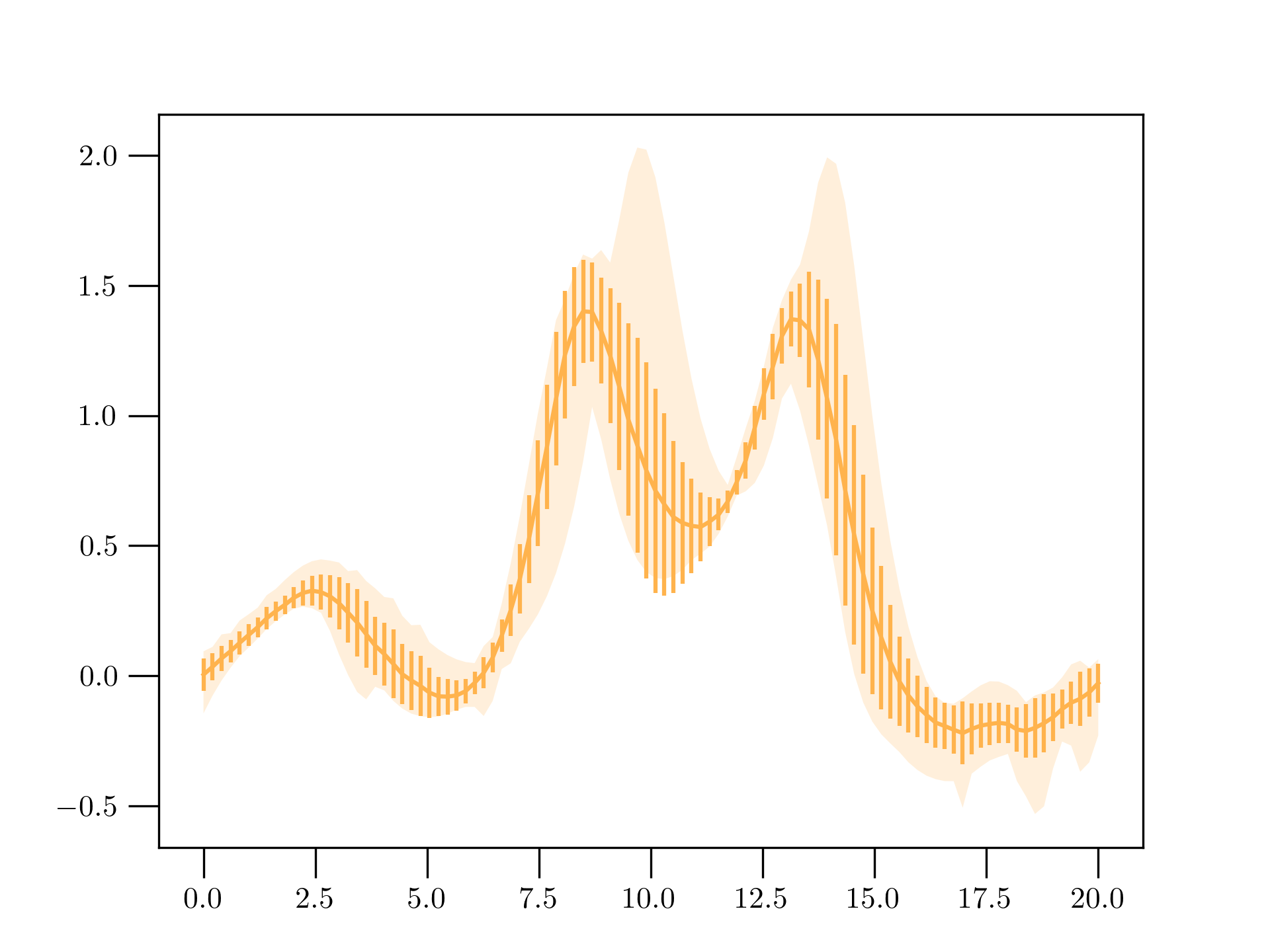}
    \caption{Baseline, 10 000 epochs, $T=1$.}
    \end{subfigure}
    \begin{subfigure}{0.48\textwidth}
    \includegraphics[width=\textwidth]{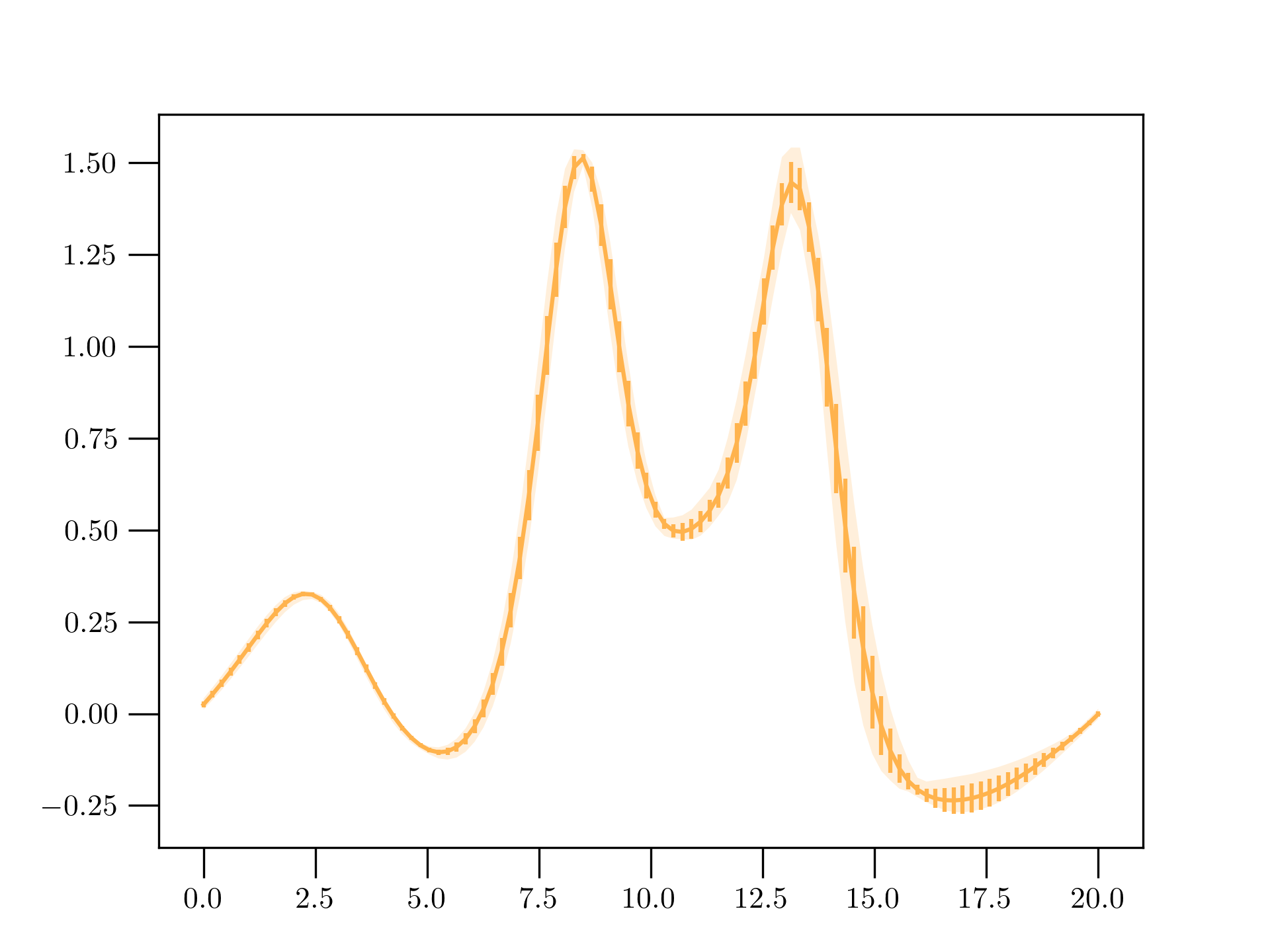}
    \caption{PHNN, 10 000 epochs, $T=1$.}
    \end{subfigure}
    \caption{Stability comparison of the baseline model and the general PHNN model. We retrain the models 20 times and compute the pointwise mean (plotted as a solid line) together with the standard deviation (plotted as error bars) and the pointwise maximum and minimum value (plotted as the shaded area).}
    \label{fig:stability_kdv_network}
\end{figure}

\begin{figure}
    \centering
    \includegraphics[width=.6\textwidth]{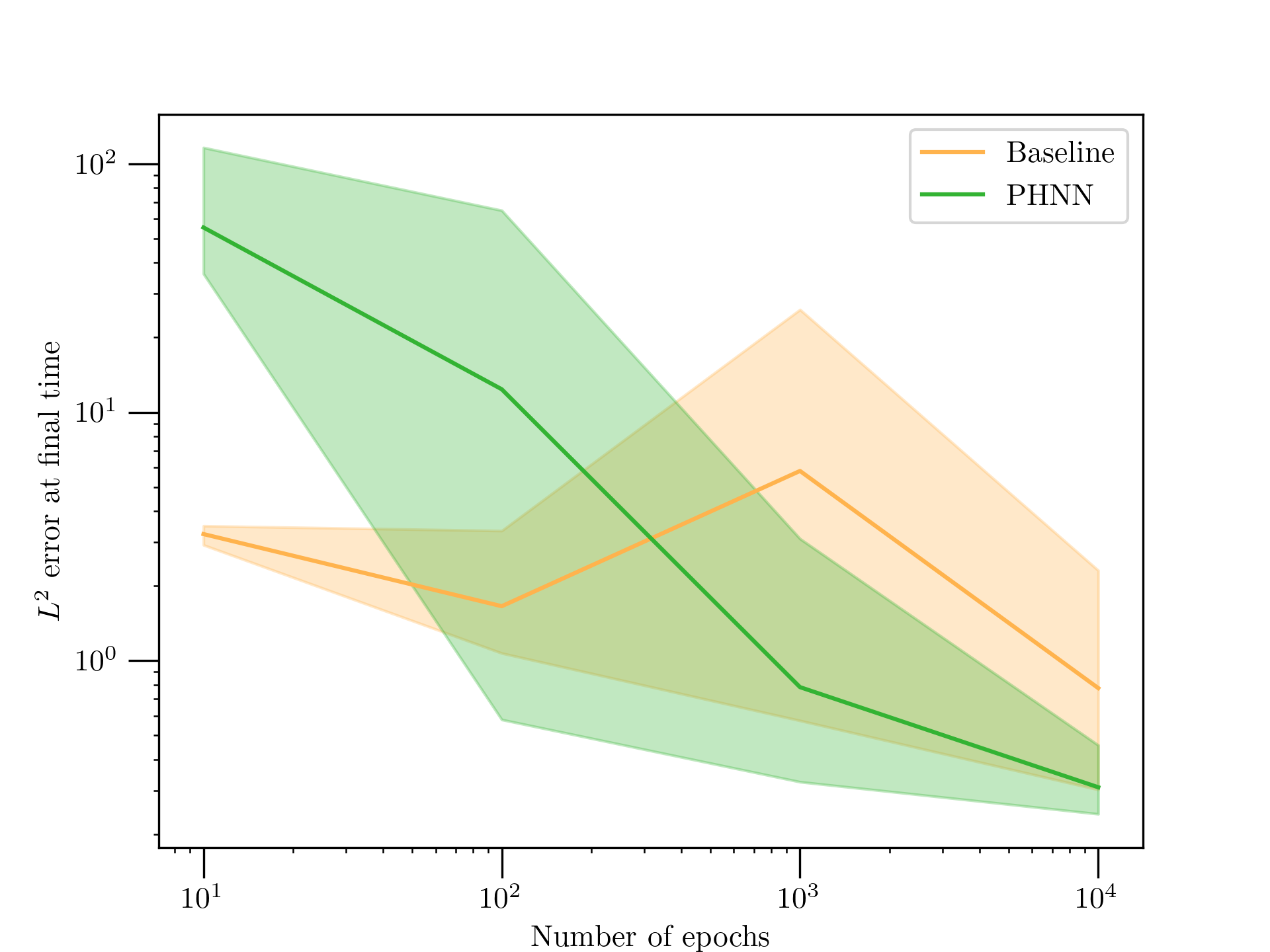}
    \caption{Convergence of the solution with respect to the number of training epochs.\label{fig:stability_epoch_convergence} Here, the shaded area represents the maximum and minimum errors obtained, and the solid middle line represents the mean value across different initial weights.}
    
\end{figure}

The training of neural networks is often observed to be quite sensitive to the intial guesses for the weights and biases of the network. Here we test this sensitivity for both the general PHNN model and the baseline model on the KdV--Burgers experiment in \Cref{sec:kdv}. We keep the training data fixed and re-generate the initial weights for the neural networks and rerun the training procedure described in algorithms \ref{alg:phnn_train} and \ref{alg:baseline_train}. In \Cref{fig:stability_kdv_network} we plot the solution at the final time together with the standard deviation and the pointwise maximum and minimum values for both the baseline model and the PHNN approach, where the standard deviation and maximum/minimum is computed across an ensemble of different initial weights for the deep neural network. In \Cref{fig:stability_epoch_convergence} we plot the $L^2$ error at the final time step against the exact solution for varying number of epochs, where the shaded areas represent the maximum and minimum values of an ensemble of varying initial weights of the neural network.

\subsection{Spatial discretization and training data}

\begin{figure}[ht!]
    \centering
    \includegraphics[width=\textwidth]{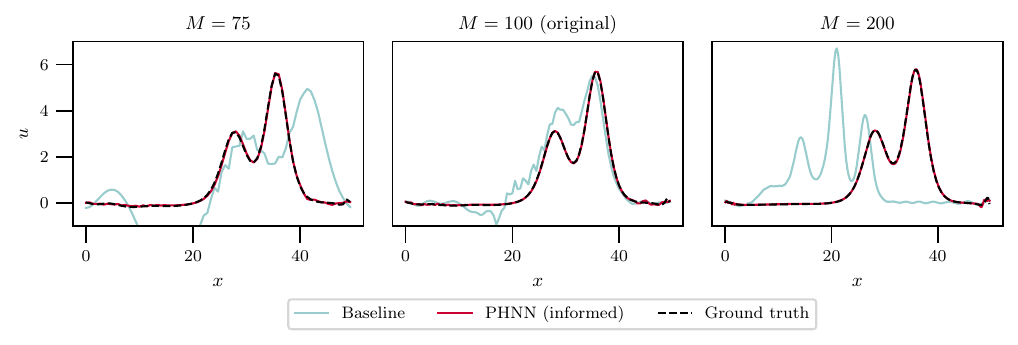}
\caption{The solution at $t=10$ obtained from models of the BBM equation \eqref{eq:bbm} with $f=0$, learned from data discretized on $M+1=101$ equidistributed points on the domain $[0,50]$. The $M$ above the plots indicates the number of equidistributed discretization points used in the integration. The PHNN model was trained for 10 000 epochs, while the baseline model was trained for 100 000 epochs, on 20 pairs of states at time $t=0$ and $t=0.4$, with initial states \eqref{eq:bbminit}.}
\label{fig:bbm_grids}
\end{figure}

We will strive to develop PHNN further to make the models discretization invariant. For now, we settle with noting that this is already a property of our model in certain cases; a sufficiently well-trained informed PHNN model will be discretization invariant if the involved integrals do not depend on derivatives. Of the examples considered in this paper, that applies to the BBM equation, the inviscid Burgers' equation, and the Cahn--Hilliard equation if $\mu=0$ in \eqref{eq:ch}. Figure \ref{fig:bbm_grids} shows how the learned BBM system can be discretized and integrated on spatial grids different from where there was training data.

\begin{figure}[ht!]
    \centering
    \includegraphics[width=\textwidth]{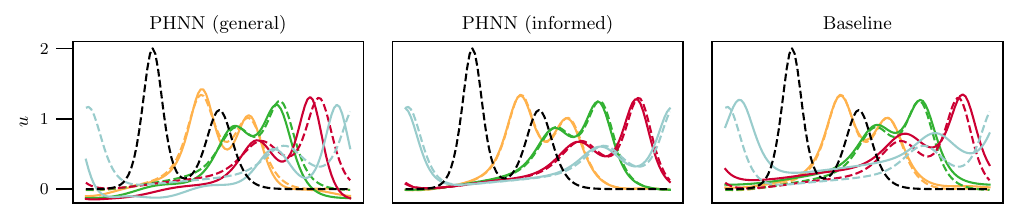}
    \includegraphics[width=\textwidth]{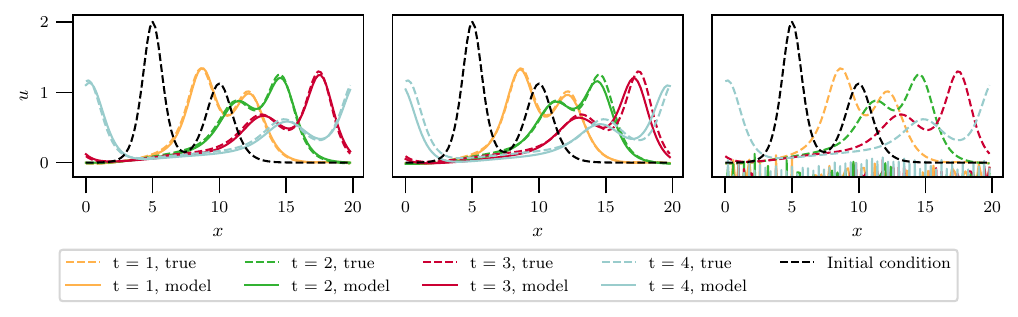}
\caption{Solution obtained from models of the KdV--Burgers equation \eqref{eq:kdvburgers}, i.e.\ without a force term, learned from 410 training states, with $10$ different initial conditions and points equidistributed in time between $t=0$ and $t=2$. \textit{Upper row:} Models trained on data generated on a spatial grid with $M=100$, same as used for training. \textit{Lower row:} Models trained on data generated on a spatial grid with $M=400$ and then downsampled to a grid with $M=100$.}
\label{fig:kdv_remesh}
\end{figure}

For the experiments in the Section \ref{sec:experiments}, we generated training data using first and second order finite difference operators to approximate the spatial derivatives. For the experiments in subsections \ref{sec:kdv} to \ref{sec:ch}, we further trained our models on the same spatial grid as the data was generated on, thus making it possible to learn convolution operators of kernel size two or three that perfectly capture the operators in the data. In a real-world scenario, this is unrealistic, as we would have to deal with discretization of a continuous system in space, as well as in time. We chose to disregard this issue in the experiments, to give a clearer comparison between PHNN and the baseline model not clouded by the error from the spatial discretization that would affect both. However, for the experiments in Section \ref{sec:kdv_compare} we tested our models on data generated on a spatial grid of four times as many discretization points, to not give the PHNNs and our baseline model an unfair advantage over the other methods. In this scenario, the data is generated from a more accurate approximation of the differential operators than what is possible to capture by the convolution operators. The results in Section \ref{sec:kdv_compare} indicate that PHNN tackles this challenge better than the baseline model, and we observe the same for the KdV--Burgers equation in Figure \ref{fig:kdv_remesh}. The PHNN models do appear to work well even with the introduction of approximation error in the spatial discretization. A more thorough study of this issue is required to gain a good understanding of how to best handle the spatial discretization.

\subsection{Sensitivity to the kernel size hyperparameter}
We note that for a new data set it is impossible to know the kernel size parameters a priori. However, in this subsection we will see that one can often distinguish feasible and infeasible kernel size parameters from simple hieristics applied to the training and validation loss. 

First we generate 400 data points from the KdV equation as described in~\Cref{sec:kdv}. We remind the reader that the kernel sizes $k=(k_1, k_2, k_3, k_4)$ are given as postive integers where the interesting values are typically $1$ or $3$. We then train the PHNN on 16 different kernel size tuples, that is we train on every possible kernel size in $\{1,3\}^4$. Following the discussion in~\Cref{sububsec:modelling_kernel_size}, we know that the feasible kernel sizes for the KdV equation are the tuples $k\in \{1,3\}^4$ such that $k_2=3$. Therefore we define the \emph{feasible} set of kernel sizes to be exactly those tuples where the second component is $3$, and the infeasible set to be the complement of this set in $\{1,3\}^4$. 

\begin{figure}
\begin{subfigure}{0.48\textwidth}
    \centering
    \includegraphics[width=\textwidth]{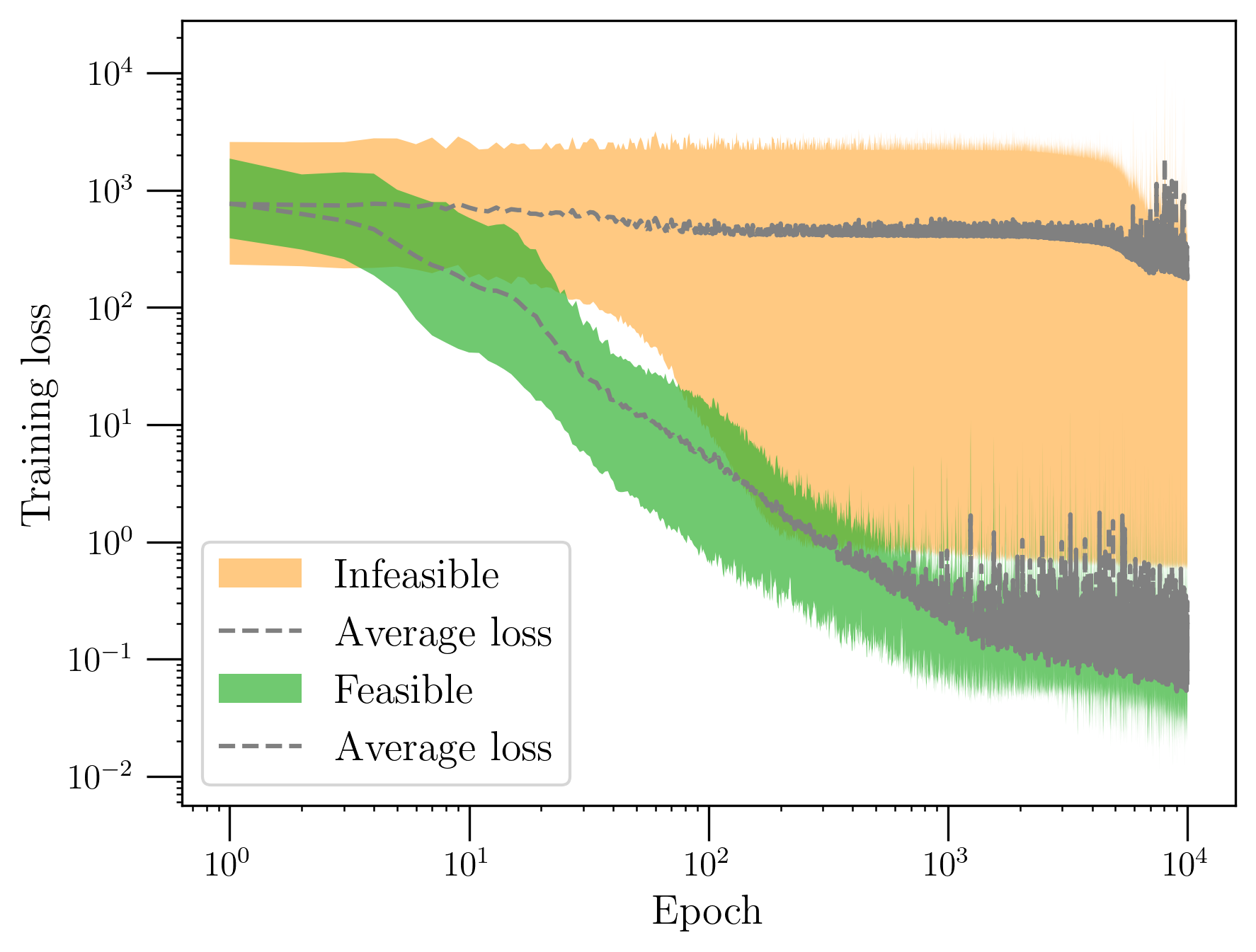}
    \caption{Training loss.}
    
\end{subfigure}
    \begin{subfigure}{0.48\textwidth}
    \centering
    \includegraphics[width=\textwidth]{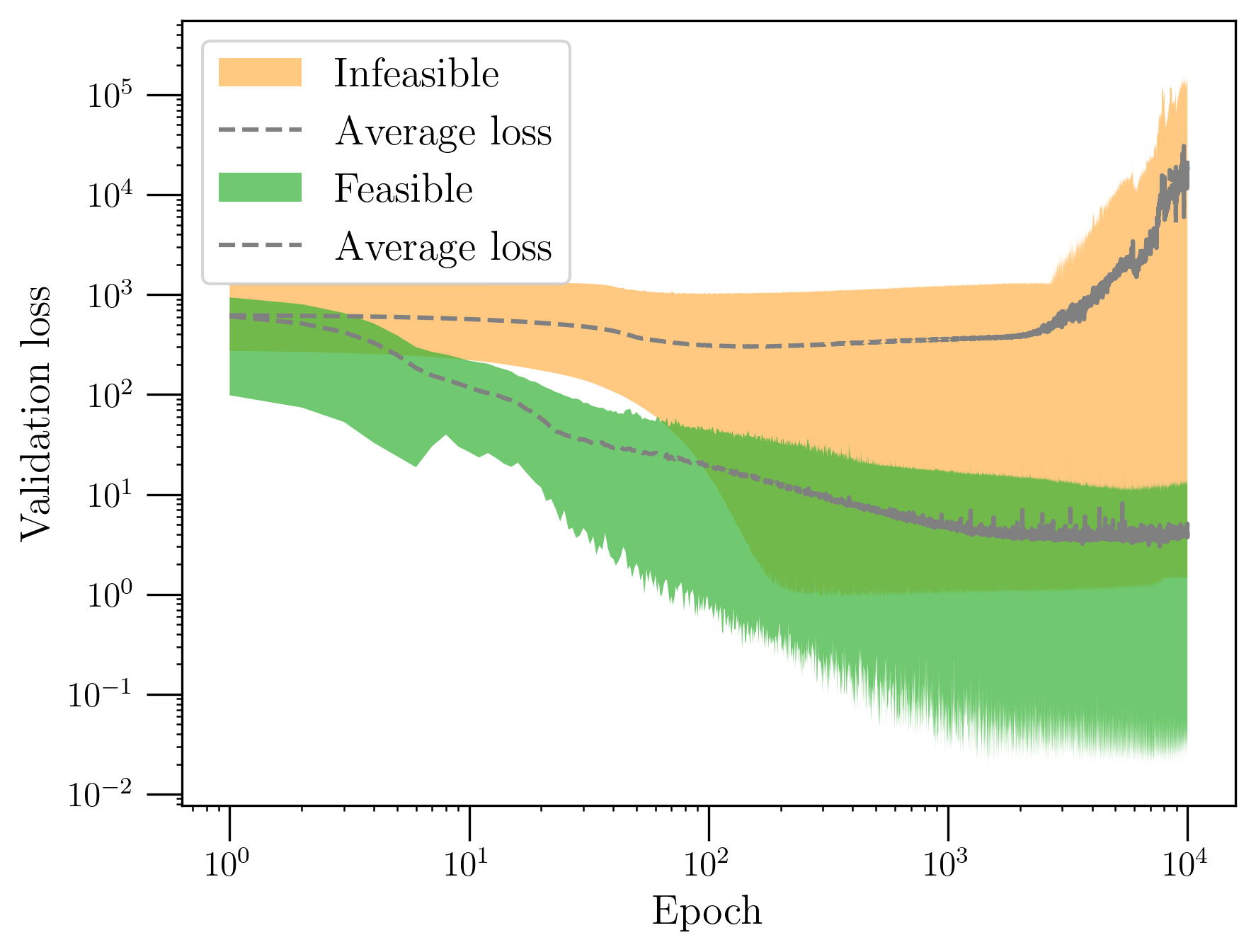}
    \caption{Validation loss.}
    \end{subfigure}
    \caption{The loss per epoch of the training data (left) and validation data (right) for different kernel sizes for the KdV equation. The shaded area represents the interval between the minimum and maximum loss for the respective kernel size set.}
    \label{fig:kernel_size_comparison}
\end{figure}

We then train on this data using 200 points for training and 200 for validation. The result is plotted in~\Cref{fig:kernel_size_comparison}. From the figure it becomes apparant that even by just looking at the training and validation data, there is a clear distinction between the feasible and infeasible kernel size hyperparameters, where infeasible kernel sizes simply do not reach convergence. We stress that this is done by only considering observation data. In other words, the training and validation loss acts as a discriminator between the feasible and infeasible kernel sizes.

An important consequence is that when encountering new data sets for which the kernel size parameter is not given, one can train on a larger set of kernel size parameters and select the ones where one does reach convergence in the training and validation loss.

\subsection{Learning more complicated skew-symmetric operators}
As we noted in Section \ref{sec:restrict}, the general pseudo-Hamiltonian formulation \eqref{eq:pdeclass} is not unique for any system. The term $f$ ensures this, but even with $f=0$ and $\mathcal{V}=0$, the integral-preserving formulation
\begin{equation}\label{eq:hamiltonianclass}
u_t = S(u^\alpha, x)\dfrac{\delta \mathcal{H}}{\delta u}[u]
\end{equation}
is not unique. For a given $\mathcal{H}$, the corresponding skew-symmetric operator is not necessarily uniquely given. Furthermore, a PDE system may have several preserved integrals. For instance, the KdV equation \eqref{eq:kdv} forms a completely integrable system, and can thus be written on the form \eqref{eq:hamiltonianclass} for infinitely many different $\mathcal{H}$ \cite{Gardner1971korteweg}. However, there are only two known \textit{Hamiltonian} formulations of the KdV equation, where $S(u^\alpha, x)$ in addition to being skew-symmetric satisfies the Jacobi identity and is called the Poisson operator \cite{Olver1993applications}. These are given by the pairs $S = \frac{\partial}{\partial x}$ and the energy functional \eqref{eq:kdvburgersH}, and $S = -\frac{1}{3} \eta(\partial_x u + u\partial_x) + \gamma^2 \partial_{xxx}$ and the momentum
\begin{equation}\label{eq:momentum}
\mathcal{H} = \frac{1}{2} \int_\Omega u^2 \, dx.
\end{equation}

Because we restricted $\hat{S}^{[k_2]}_\theta$ to be constant in our models in Section \ref{sec:experiments}, we achieved uniqueness and learned the formulation where $\mathcal{H}$ represents the energy. To learn the formulation where $\mathcal{H}$ represents momentum, we would have to let $\hat{S}^{[k_2]}_\theta$ depend on $u$, and we would need $k_2 \geq 5$ to learn the third spatial derivative. Furthermore, we would need to restrict the kernel of the convolutional layer in $\hat{\mathcal{H}}_\theta$ to be of size $1$, so that this could not learn the energy functional. That is, the alternative Hamiltonian formulation could be learned by restricting $\hat{\mathcal{H}}_\theta$ more and $\hat{S}^{[k_2]}_\theta$ less. An exploration of using our models to learn alternative pseudo-Hamiltonian formulations of the same system is a planned future direction.

Such an exploration will also involve considering more PDE systems. One interesting candidate is the modified Korteweg--de Vries (mKdV) equation \cite{Miura1968korteweg}
\begin{equation}
u_t + \eta u^2 u_x - \gamma^2 u_{xxx} = 0,
\end{equation}
which introduces a more complicated $S$ yet. For the momentum \eqref{eq:momentum}, the corresponding Poisson operator is in this case $S = -\frac{2}{3} \eta \partial_x u \partial_x^{-1} u \partial_x + \gamma^2 \partial_{xxx}$, and thus includes both derivatives, an antiderivative and the system state $u$ \cite{Wang2002recent}. Setting $k_2 = M$ and letting $\hat{S}^{[M]}_\theta$ depend on $u$, it would be expressive enough to learn a consistent discretization of this $S$. How to impose uniqueness on this formulation is not trivial, and one of the things we will investigate in future work.

\subsection{Proof of convergence in the idealized case}
In this section we show a simplified error estimate for learning the right hand side of an ODE. Consider thus a model ODE of the form

\begin{equation}
\label{eq:modelode}
    \begin{cases}
    \dot{u}(t)&=g(u(t), t)\\
    u(0)&=u_0,
    \end{cases}
\end{equation}
where $u:[0,T)\to\R^M$ and $g:\R^M\to \R^M$. Note that the spatially discretized equation \eqref{eq:odesys} can be cast into this form. 
In a certain sense, both the baseline model and the PHNN model tries to identify $g$ by minimizing the $L^p$-norm of the observations $u(t^j)$ and the predictions $u_\theta(t^j)$. On a high level, this gives us a sequence $u_\theta\to u$ in $L^p$, but as is well-known, this would not be enough to conclude anything about the convergence of $g_\theta\to g$, since $L^p$ convergence in general does not imply convergence of the the derivatives. However, by utilizing the fact that we have a certain control over the discretized temporal derivatives in the learning phase, we can show that the $g_\theta\to g$ in the same $L^p$ norm, \emph{provided the training loss is small enough}. The following theorem makes this precise.

\begin{theorem}
Let $\Delta t>0$, and $g, \tilde{g}:\R^M\to \R^M$. Assume that $u:[0, T)\to \R^M$ solves \eqref{eq:modelode} and that $\tilde{u}^1, \ldots, \tilde{u}^N\in \R^M$ obey\footnote{This is essentially saying they are obtained during training for the loss function.}
\begin{equation}
    \label{eq:forward_euler_dnn}
    \frac{\tilde{u}^{j+1}-u^j}{\Delta t} = \tilde{g}(u^j)\qquad \text{for }j=0, \ldots, N-1.
\end{equation}
Then, 
\[\left(\Delta t \sum_{j=1}^{N-1}\left(g(u(t^j), t^j)-\tilde{g}(u^j, t^j)\right)^p\right)^{1/p}\leq\frac{1}{\Delta t} \left(\sum_{j=1}^N \Delta t \left|u^j-\tilde{u}^{j}\right|^p\right)^{1/p} + C_g\Delta t.\]
\end{theorem}
\begin{proof}
Define $u^0, \ldots, u^N\in \R^M$ as
\begin{equation}
\label{eq:forward_euler}    
u^j:=u(t^n)  \qquad j  =1, \ldots, N.
\end{equation}
By a Taylor expansion, we have 
\[\frac{u^{j+1}-u^j}{\Delta t} = g(u^n) + \left(\left[\frac{\partial g}{\partial u}(u(\xi), \xi)\right]g(u(\xi, \xi)) + \frac{\partial g}{\partial t}(u(\xi), \xi)\right)\Delta t\qquad \xi \in [t^j, t^{j+1}], j=0, \ldots, N-1.\]
Hence, we get
\[\left(\sum_{j=0}^{N-1} \Delta t \left|g(u(t^j), t^j)-\tilde{g}(u^j, t^j)\right|^p\right)^{1/p}\leq\left(\sum_{j=1}^N \Delta t \left|\frac{u^j-u^{j-1}}{\Delta t}-\frac{\tilde{u}^j-u^{j-1}}{\Delta t}\right|^p\right)^{1/p} + C_g\Delta t. \]
We furthermore have
\begin{align*}
 \left(\sum_{j=1}^N \Delta t \left|\frac{u^j-u^{j-1}}{\Delta t}-\frac{\tilde{u}^j-u^{j-1}}{\Delta t}\right|^p\right)^{1/p}
&=\left(\sum_{j=1}^N \Delta t \left|\frac{u^j-\tilde{u}^{j}}{\Delta t}-\frac{u^{j-1}-u^{j-1}}{\Delta t}\right|^p\right)^{1/p}\\
&= \left(\sum_{j=1}^N \Delta t \left|\frac{u^j-\tilde{u}^{j}}{\Delta t}\right|^p\right)^{1/p}\\
&= \frac{1}{\Delta t} \left(\sum_{j=1}^N \Delta t \left|u^j-\tilde{u}^{j}\right|^p\right)^{1/p}.\qedhere
\end{align*}
\end{proof}
\begin{remark}
    We note that \eqref{eq:forward_euler_dnn} means that the sequence $\{\tilde{u}^j\}_j$ is the learning data obtained by either the baseline or PHNN algorithm using the forward Euler integrator.
\end{remark}
\begin{remark}
In the above theorem, $\left(\sum_{i=1}^N \Delta t \left|u^j-\tilde{u}^{i}\right|^p\right)^{1/p}$ is proportional to the training loss. In other words, the error in the approximation of $g$ we get is bounded by $1/\Delta t \cdot (\text{Training loss})$. Hence, to achieve an accuracy $\epsilon$ in $g$, we need to train to a training loss of $\epsilon\Delta t$.
\end{remark}

\section{Conclusions}

One of the advantages of PHNN is that it facilitates incorporating prior knowledge and assumptions into the models. The advantage of this is evident from the experiments in Section \ref{sec:experiments}; the informed PHNN model performs consistently very well. We envision that our models can be used in an iterative process where you start by the most general model and as you learn more about the system from this, priors can be imposed, eventually resulting in models that share certain geometric properties with the underlying system.

As discussed in Section \ref{sec:stability}, the PHNN models are highly sensitive to variations in the initialized parameters of the neural networks. By the numerical results, the general PHNN model in particular seems to be more sensitive to this than the baseline model. However, the best trained PHNN models outperform the baseline models across the board. In a practical setting, where the ground truth is missing, we could train a number of models with different initialization of the neural networks and disregard those that deviate greatly from the others.

The aim of this paper has been to introduce a new method, demonstrate some of its advantages and share the code for the interested reader to study and develop further. It is the intent of the authors to also continue the work on these models. For one, we will be doing further analysis to address the issues raised in the previous section and improve the training of the models under various conditions. Secondly, we would like to extend the code to also work on higher-dimensional PDEs, and consider more advanced problems. One of the most promising uses of the methodology may be on image denoising and inpainting, motivated by the results of sections \ref{sec:pm} and \ref{sec:ch}. Lastly, the pseudo-Hamiltonian formulation could be used with other machine learning models than neural networks. Building on \cite{Holmsen2023pseudo}, we will develop methods for identifying analytic terms for one or several or all of the parts of the pseudo--Hamiltonian model \eqref{eq:phnn}, and compare the performance to existing system identification methods like \cite{Rudy2017data, Schaeffer2017learning, Kaheman2020sindy}. We are especially intrigued by the possibility to identify the integrals of \eqref{eq:pdeclassspef}, while the external forces might be best modelled by a neural network.

\subsection*{Acknowledgement}
This work was supported by the research project PRAI (Prediction of Riser-response by Artificial Intelligence) financed by the Research Council of Norway, project no.\ 308832, with Equinor, BP, Subsea7, Kongsberg Maritime and Aker Solutions. The authors are grateful to Brynjulf Owren for illuminating discussions, and to the anonymous reviewers for their insightful comments and suggestions. Furthermore, the authors thank Katarzyna Michałowska and Signe Riemer-Sørensen for helpful comments on the manuscript, Eivind Bøhn for help with coding issues, and Benjamin Tapley for both.

\bibliographystyle{amsplain}
\bibliography{phnn_pde_bib}
\end{document}